%% file: main.tex
\documentclass[10pt,twocolumn,letterpaper]{article}

\usepackage{cvpr}              
\usepackage[table]{xcolor}

\usepackage{graphicx}
\usepackage{float} 
\usepackage[accsupp]{axessibility}

\definecolor{cvprblue}{rgb}{0.21,0.49,0.74}
\usepackage[pagebackref,breaklinks,colorlinks,allcolors=cvprblue]{hyperref}
\usepackage{subcaption}

\usepackage[accsupp]{axessibility}  

\usepackage{booktabs}
\usepackage{threeparttable}
\usepackage{multirow}

\usepackage{booktabs}
\usepackage{pgfplots}
\usepackage{tikz}
\usepackage{xcolor}
\pgfplotsset{compat=1.18}

\definecolor{blueChart}{RGB}{55, 138, 221}
\definecolor{coralChart}{RGB}{216, 90, 48}
\definecolor{grayChart}{RGB}{136, 135, 128}

\def\paperID{18} 
\def\confName{CVPR}
\def\confYear{2026}

\title{Horse Eye Blink Detection and Classification for Equine Affective State Assessment}

\author{
    \begin{tabular}{cccc}
        João Alves$^{1}$ & Signe Møller-Skuldbøl$^{1}$ & Pia Haubro Andersen$^{2}$ & Rikke Gade$^{1}$ \\
        {\tt\small jmal@create.aau.dk} & {\tt\small smolle22@student.aau.dk} & {\tt\small pia.haubro.andersen@slu.se} & {\tt\small rg@create.aau.dk} \\
    \end{tabular} \\[0.3cm]
    \small
    $^{1}$Visual Analysis and Perception Lab, Aalborg University \\
    \small
    $^{2}$Department of Animal Biosciences, Swedish University of Agricultural Sciences
}

\begin{document}
\maketitle
\input{sections/0abstract}    
\input{sections/1intro}
\input{sections/2background}

\input{sections/3Dataset_Methods}

\input{sections/4Experiments_Results}

\input{sections/5Discussion_Conclusion}

\clearpage
{
    \small
    \bibliographystyle{unsrt}
    \bibliography{bibliography}
}
\input{sections/suppl}


\end{document}

%% file: sections/0abstract.tex
\begin{abstract}
Automated detection of equine facial action units (AUs) is a promising yet under-explored avenue for pain and affective state assessment in horses. Half and full-blink movements are recognised indicators of pain and stress, but as micro-expressions, their subtle, fine-grained nature makes them easily missed by the naked eye and only discernible through frame-by-frame video inspection, making reliable automated detection from video a particularly demanding task. We develop and evaluate three methods for automated blink classification from horse videos: a frame-based YOLOv12 detector, an optical flow magnitude thresholding approach, and a fine-tuned VideoMAE model, tested on a publicly available dataset. We achieve a macro-F1 score of 0.898 when doing blink classification and 0.926 on binary blink detection. Our results highlight both the potential and the inherent challenges of fine-grained AU detection for equine welfare monitoring.
\end{abstract}

%% file: sections/1Intro.tex
\section{Introduction}
\begin{figure}
    \centering
    \includegraphics[width=\linewidth]{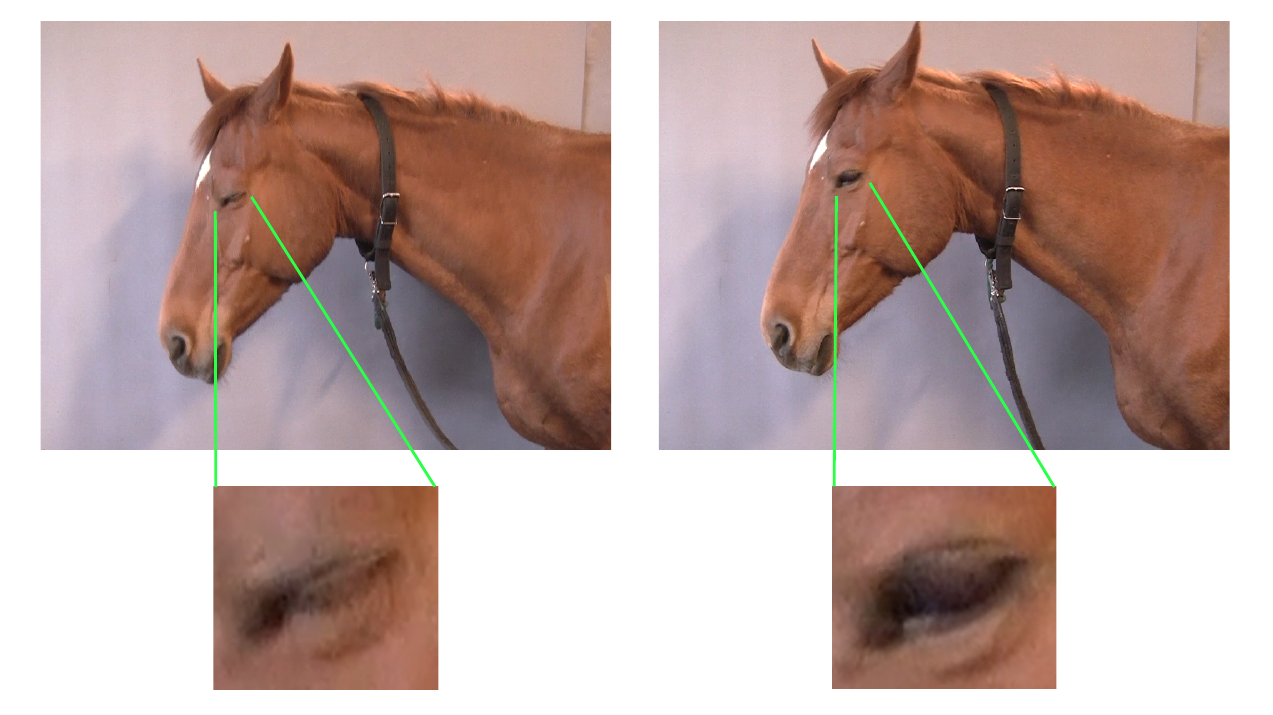}
    \caption{Example eye blink movements~\cite{rashidEquineFacialAction2020} data: full-blink (left) and half-blink (right).}
    \label{fig:blinks}
\end{figure}

Research on human pain assessment generally benefits from the use of self-reporting, whereas in the animal domain, it is not available. Observational scales such as the Horse Grimace Scale (HGS)~\cite{dallacostaDevelopmentHorseGrimace2014} are used in clinical practice, while other research relies on the more objective Equine Facial Action Coding System (EquiFACS)~\cite{wathanEquiFACSEquineFacial2015}, which builds on the human centered FACS~\cite{ekman2002}. EquiFACS describes facial movements based on the animal's musculature~\cite{broomeGoingDeeperTracking2023}, but producing these annotations is a resource demanding task \cite{andersenMachineRecognitionFacial2021}. Past works have shown that specific facial action units and their co-occurrence can reveal an animal's emotional state. In particular, shifts in blink-related action units frequency have been linked to pain and stress in horses~\cite{dallacostaDevelopmentHorseGrimace2014, rashidEquineFacialAction2020, lundbladExploringFacialExpressions2024}, therefore, this work focuses on the detection and classification of full-blink and half-blink movements (see Figure \ref{fig:blinks}) from horse video data to enable automated action unit detection for affective state assessment. We propose and evaluate multiple methodologies on a publicly available dataset~\cite{rashidEquineFacialAction2020}. Our contributions are as follows:

\begin{itemize}
    \item A human baseline study demonstrating the perceptual difficulty of equine full- and half-blink annotation, showing that even naive human observers struggle with half-blink classification ($73\%$ agreement, Krippendorff's $\alpha=0.57$).
    \item Development of three methods for equine blink detection and classification: an optical flow magnitude binary thresholding method, a frame-based YOLOv12 approach, and a fine-tuned VideoMAE model.
    \item Evaluation  of our methods on the publicly available set~\cite{gleerupEquinePainFace2015, rashidEquineFacialAction2020}, covering both temporal blink event detection and fine-grained full vs.\ half-blink classification.
\end{itemize}

%% file: sections/2background.tex
\section{Background}

Rashid et al.~\cite{rashidEquineFacialAction2020} studied EquiFACS in horses under experimental and clinical pain, identifying ear rotator movements, nostril dilation, and lower-face behaviours as key indicators. Ask et al. ~\cite{askChangesEquineFacial2024} further showed that AUs in the ear, eye, and lower-face regions increased with orthopaedic pain severity, framing pain as a dynamic process reflected in changing expressions. Andersen et al.~\cite{andersenMachineRecognitionFacial2021} demonstrated the feasibility of automated equine pain detection by combining EquiFACS with machine learning and cross-domain morphing of animal features, while highlighting the high annotation cost. Further works have attempted automated EquiFACS AU recognition with varying AU selections~\cite{liAutomatedDetectionEquine2021, alvesReadMyEars2025}. In Dimitrova et al. ~\cite{panoutsos_automatic_2024}, the authors specifically targeted full-blink detection in horse videos using classical trackers and a neural network, but encountered several limitations and did not address different blink types. In the human domain, blink detection has been tackled with diverse architectures, including convolutional networks~\cite{delacruzEyeLRCNLongTermRecurrent2024}, transformer-based models~\cite{fodorBlinkLinMulTTransformerBasedEye2023}, and dual-embedding video vision transformers~\cite{hongRobustEyeBlink2024}. This research is supported by numerous benchmark datasets such as CEW~\cite{songEyesClosenessDetection2014}, ZJU~\cite{panEyeblinkbasedAntiSpoofingFace2007}, MRL Eye~\cite{fusekPupilLocalizationUsing2018}, RT-BENE~\cite{cortaceroRTBENEDatasetBaselines2019}, and EyeBlink8 and Researcher's Night~\cite{fogeltonEyeBlinkCompleteness2018}, which constitutes a major advantage over the equine domain, where public datasets are scarce. While some methods classify incomplete or partial blinks~\cite{delacruzEyeLRCNLongTermRecurrent2024}, performance on this task remains considerably lower than for full-blink detection. The primary obstacle to EquiFACS detection studies is the scarcity of available training data~\cite{andersenMachineRecognitionFacial2021}. A key contributing factor to this is the time-intensive nature of EquiFACS annotation, which requires trained annotators to manually review long video sequences in their entirety. To mitigate the bottleneck imposed by exhaustive manual annotation in equine pain research, we consider the creation of semi or fully automated annotation tools. By narrowing the annotation search space, such tools could substantially reduce the time required to build large annotated datasets, facilitating the development of more capable equine pain assessment systems. This work investigates multiple approaches to the equine eye blink detection problem, with the goal of streamlining annotation and reducing EquiFACS dataset production cost.

%% file: sections/3Dataset_Methods.tex
\section{Methodology}
We consider two tasks related to EquiFACS annotation process. First, the task of automated blink detection (both half or full-blinks) in video: given an input video, the goal is to identify temporal regions of interest where blink activity occurs. The second task concerns the fine-grained classification of blink activity in horses. Given either an individual frame or a short video clip, the goal is to categorise the eye state into one of three classes: \textit{none} (open eye), \textit{half-blink}, or \textit{full-blink} (see Figure \ref{fig:blinks}). This is a non-trivial problem due to the subtle differences in eyelid tension, and motion-induced distortion. For our studies, we consider the data from ~\cite{gleerupEquinePainFace2015}, which consists of videos of 6 horses from different breeds recorded at 1080p@25 FPS. A publicly available subset of 12 videos (S1-S12, with 2 videos per horse)~\cite{rashidEquineFacialAction2020} provides expert EquiFACS annotations for each video, a comprehensive resource for AU analysis. We use this public subset as our test set to ensure result reproducibility. The remainder of the data, drawn from the private portion of ~\cite{gleerupEquinePainFace2015}, was annotated as part of this work to construct the train and validation sets, with strict separation to prevent data leakage. To accommodate the two categories of methods evaluated in this work, the data was annotated in two formats. For frame-based methods, individual frames were labelled according to eyelid closure. Class \textit{half-blink} was assigned when the eye was at least halfway closed, with subtle blinks and minor eye tension ignored to reduce sensitivity to variation in horse eye shapes. The \textit{full-blink} class required the eye to be fully closed with no eyeball visible; consequently, any eye closed more than halfway but with the eyeball still visible was labelled as \textit{half-blink}. For video-based methods, videos from the train set were manually annotated via EquiFACS typed annotations, and sampled into single class video clips. At test time, both method types are evaluated on data originating from the public test set~\cite{rashidEquineFacialAction2020}, both on frame-level labels and individually classified short video clips. In our experiments, data augmentation was applied to the train data, including horizontal flipping as well as hue, brightness, saturation and contrast jittering.

\subsection{YOLO fine-tuning} \label{sec:yolo}

We train a YOLOv12n detector for per-frame eye closure classification into three classes: \textit{none}, \textit{half-blink}, and \textit{full-blink}, trained on 13,206 manually annotated frames from 6 horses (77.0\% open, 13.6\% half-blink, 9.4\% full-blink), using leave-one-out cross-validation. Since the model operates frame-by-frame, this approach discards all temporal dynamics and two post-processing steps are applied to produce temporally coherent predictions. First, we assign a single label to all consecutive blink frames within an episode, with \textit{full-blink} taking precedence over \textit{half-blink} when both are present. Second, asymmetric padding extends each detected episode by a fixed number of frames on either side, compensating for the conservative \textit{half-blink} threshold; the padding is applied to any blink episodes and padding duration is determined via hyperparameter search.

\subsection{Optical flow magnitude thresholding}

Our optical flow method works by leverages motion information directly. It reuses the YOLO eye detector from Section \ref{sec:yolo} to extract per-frame bounding boxes, which are smoothed over time to reduce jitter in the box centre, width, and height. The cropped eye regions are resized to a common dimension, converted to greyscale, and dense optical flow~\cite{farnebackTwoFrameMotionEstimation2003} is computed to yield a mean motion magnitude per frame. A blink threshold is set at the midpoint of the videos's magnitude range, with fixed padding applied around each detected episode. Unlike the YOLO approach, this method is not suited for real-time analysis and is limited to binary blink detection, with no distinction between \textit{half-} and \textit{full-blink}. This classical method provides a point of reference for the learning-based approaches.

\subsection{VideoMAE fine-tuning} \label{sec:videomae}
VideoMAE~\cite{tongVideoMAEMaskedAutoencoders2022} has proven to be an efficient feature extractor for action recognition tasks. We fine-tune VideoMAE-Base, a masked autoencoder pre-trained on the Kinetics-400 dataset, for 3-class eye blink event video clip classification (\textit{none}, \textit{half-blink}, \textit{full-blink}) using cross-entropy loss and leave-one-out validation. The model processes video frames as 16×16 patch tokens across 16 temporal frames. We replace the original 400-class head with a 3-class linear classifier head. Prior to feeding frames into VideoMAE, the YOLO eye detector from Section \ref{sec:yolo} is used again to crop to the eye region. All videos are normalised to exactly 16 frames through frame repetition or downsampling, ensuring consistency with VideoMAE's fixed position embeddings.

%% file: sections/4Experiments_Results.tex
\begin{figure}
    \centering
    \begin{subfigure}{0.49\linewidth}
        \centering
        \includegraphics[width=\linewidth]{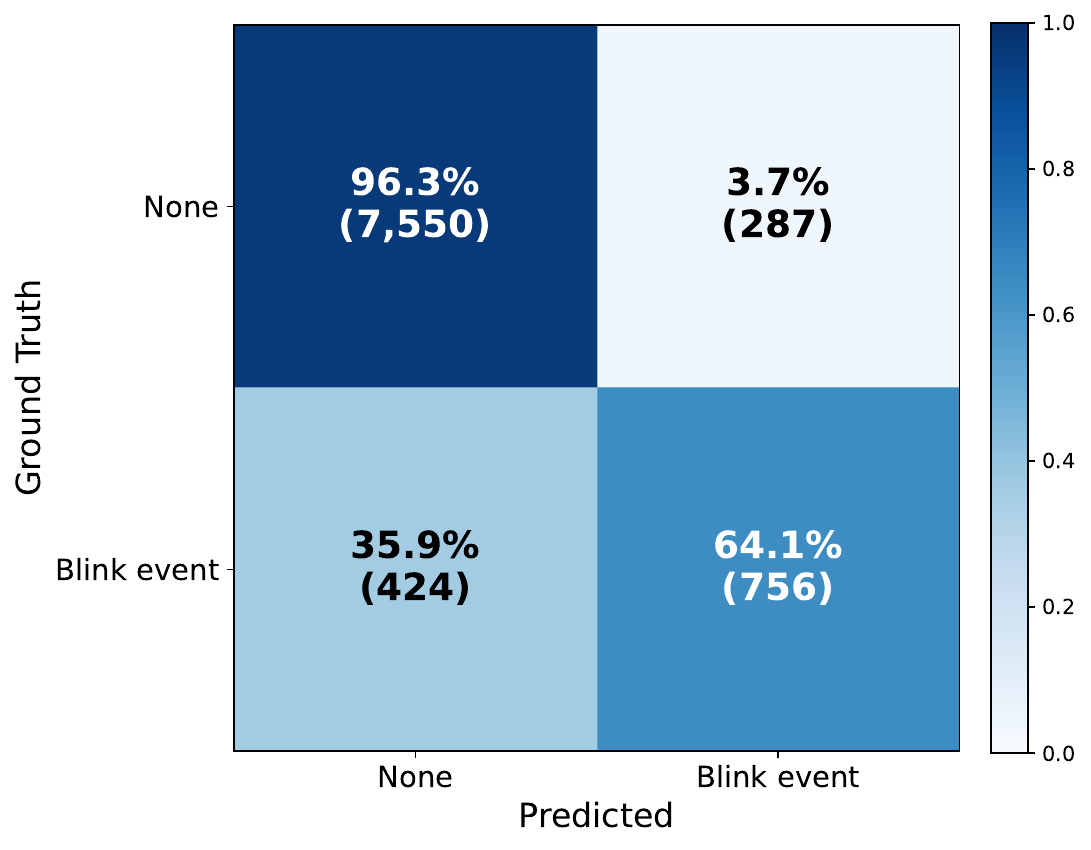}
        \caption{Optical flow (frames)}
\end{subfigure}
    \begin{subfigure}{0.49\linewidth}
        \centering
        \includegraphics[width=\linewidth]{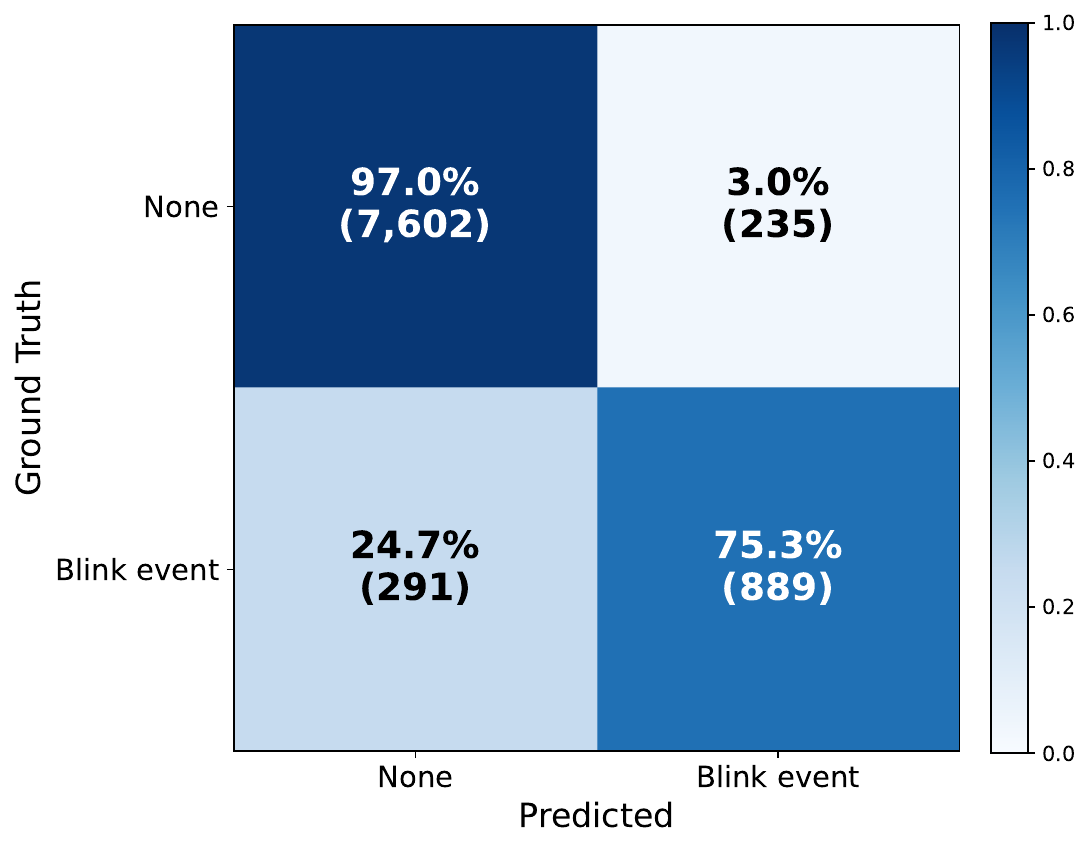}
        \caption{YOLO (frames)}
    \end{subfigure}
    \begin{subfigure}{0.49\linewidth}
        \centering
        \includegraphics[width=\linewidth]{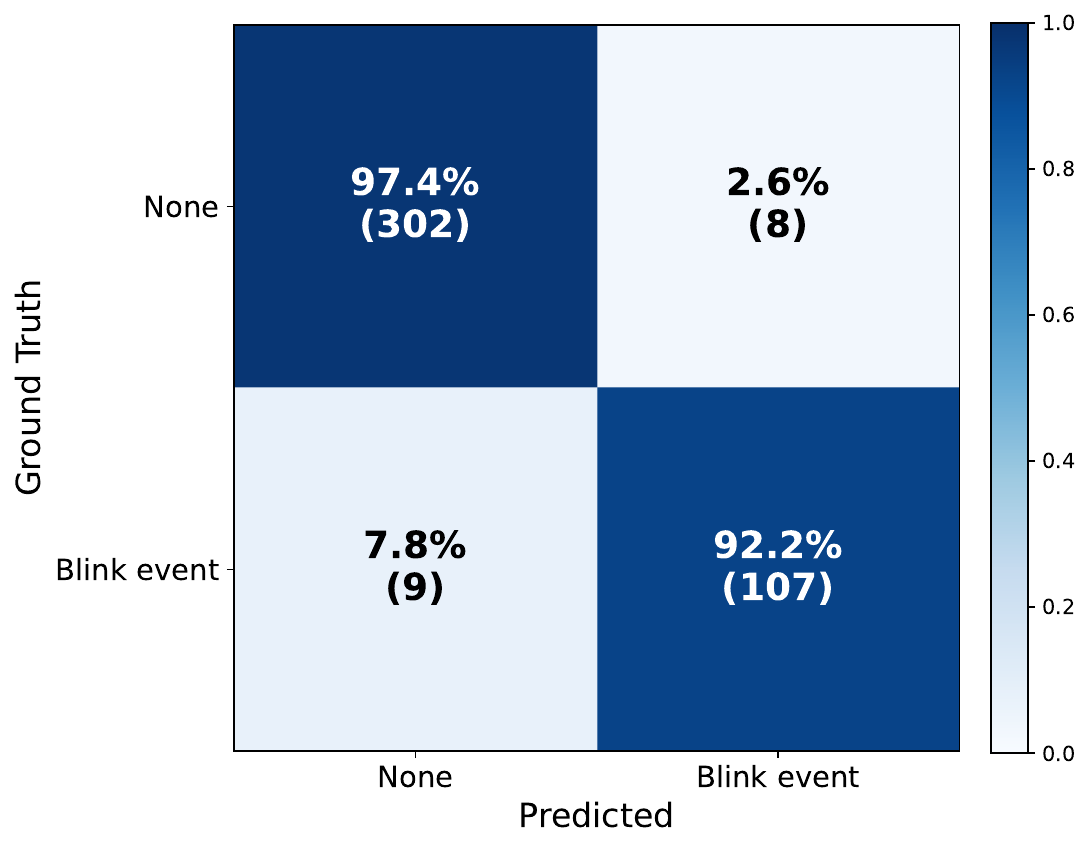}
        \caption{YOLO (video)}
    \end{subfigure}
    \begin{subfigure}{0.49\linewidth}
        \centering
        \includegraphics[width=\linewidth]{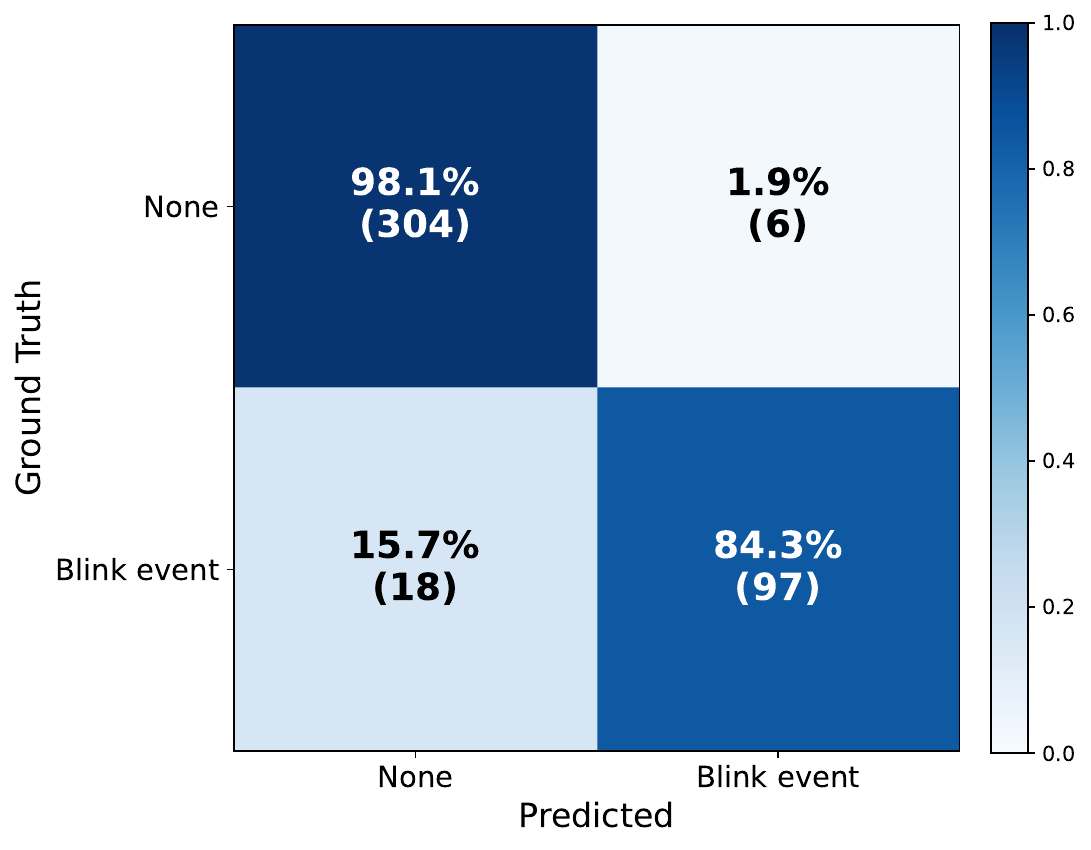}
        \caption{VideoMAE (video)}
    \end{subfigure}
    \caption{Confusion matrices for test set evaluation for each method on blink event detection task, with note on test set used (frame or video based).}
    \label{fig:CM_det}
\end{figure}

\begin{figure}
    \centering
    \begin{subfigure}{0.49\linewidth}
        \centering
        \includegraphics[width=\linewidth]{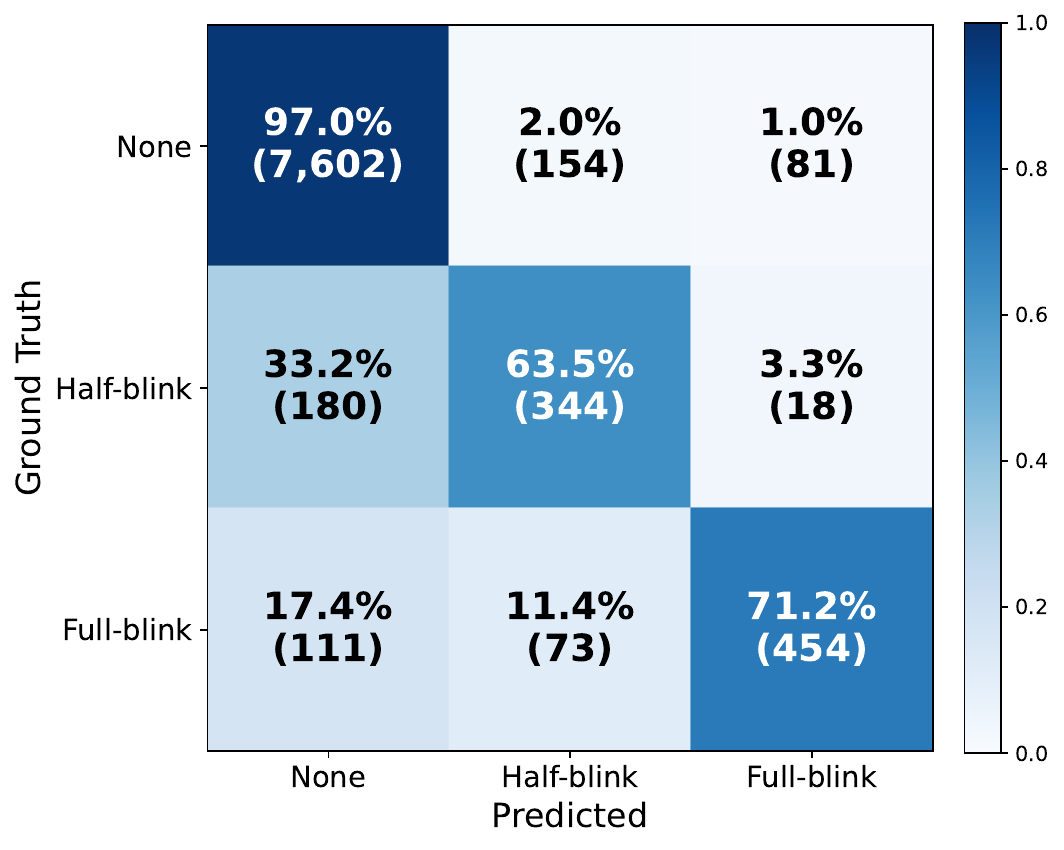}
        \caption{YOLO (frames)}
    \end{subfigure}
    \begin{subfigure}{0.49\linewidth}
        \centering
        \includegraphics[width=\linewidth]{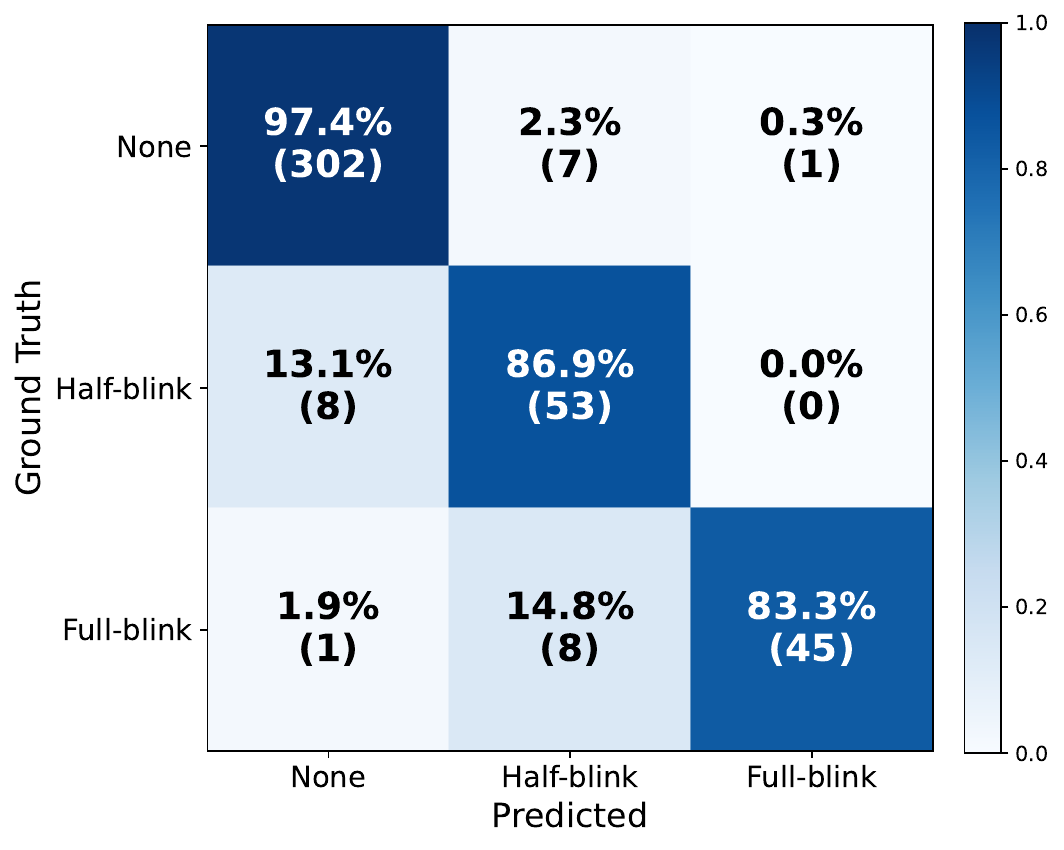}
        \caption{YOLO (video)}
    \end{subfigure}
    \begin{subfigure}{0.49\linewidth}
        \centering
        \includegraphics[width=\linewidth]{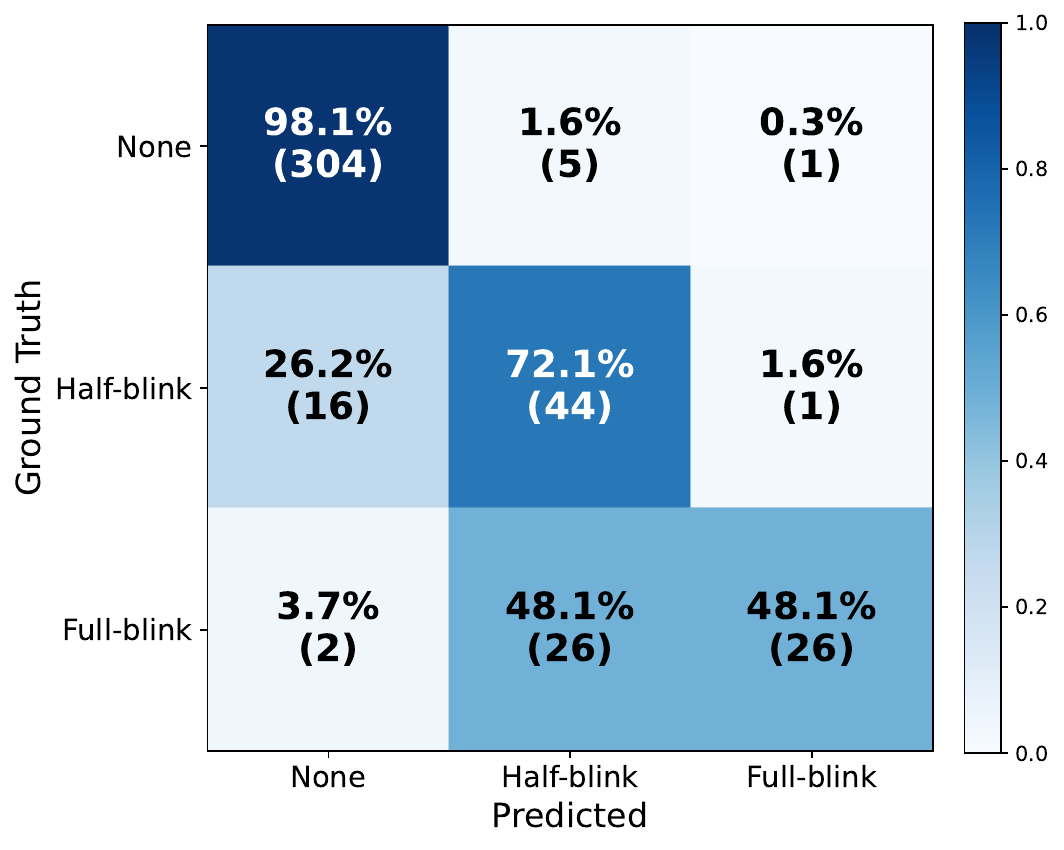}
        \caption{VideoMAE (video)}
    \end{subfigure}
    \caption{Confusion matrices for test set evaluation for each method on blink event classification task, with note on test set used (frame or video based).}
    \label{fig:CM_class}
\end{figure}

\begin{figure}
    \centering
    \includegraphics[width=0.65\columnwidth]{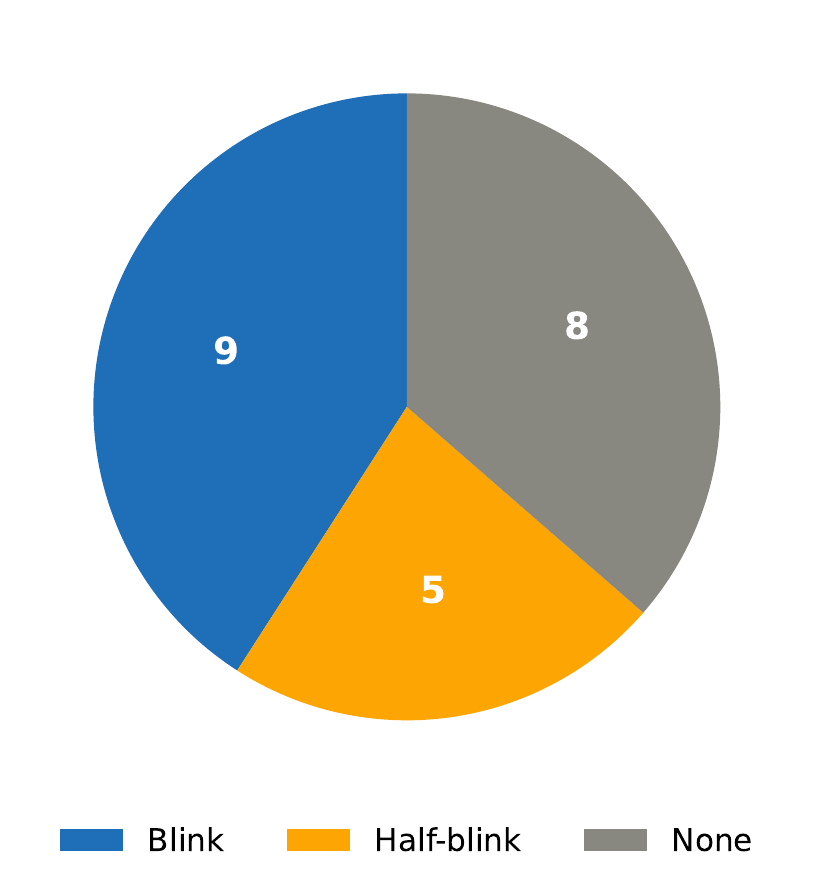}
    \caption{Inter-annotator study data class distribution with 22 total clips sampled from the video test set.}
    \label{fig:blink_dist}
\end{figure}

\begin{table*}
    \footnotesize
    \centering
    \begin{tabular}{llccccccccc}
        \toprule
        \multirow{2}{*}{\textbf{Method}} &
        \multirow{2}{*}{\textbf{Test set}} &
        \multirow{2}{*}{\textbf{Accuracy}} &
        \multirow{2}{*}{\textbf{Macro Precision}} &
        \multirow{2}{*}{\textbf{Macro Recall}} &
        \multirow{2}{*}{\textbf{Macro F1}} &
        \multicolumn{3}{c}{\textbf{Per-Class F1}} \\
        \cmidrule(lr){7-9}
        & & & & & & \textit{None} & \textit{Half-blink} & \textit{Full-blink} \\
        \midrule
        \multicolumn{9}{c}{\textit{Binary detection (blink event / none)}} \\
        \midrule
        Optical Flow ($\tau$=0.5) & Frame & 0.921 & 0.725 & 0.641 & 0.680 &  - & - & - \\
        YOLOv12                   & Frame & 0.942 & 0.877 & 0.862 & 0.870 & - & - & - \\
        \midrule
        YOLOv12                   & Video & \textbf{0.960} & 0.930 & \textbf{0.922} & \textbf{0.926} & - & - & -\\
        VideoMAE                  & Video & 0.944 & \textbf{0.943} & 0.912 & \textbf{0.926} & - & - & - \\
        \midrule
        \multicolumn{9}{c}{\textit{Three-class classification (none / half-blink / full-blink)}} \\
        \midrule
        YOLOv12                   & Frame & 0.932 & 0.796 & 0.772 & 0.782 & 0.967 & 0.618 & 0.763 \\
        \midrule
        YOLOv12                   & Video & \textbf{0.941} & \textbf{0.910} & \textbf{0.892} & \textbf{0.898} & \textbf{0.973} & \textbf{0.822} & \textbf{0.900} \\
        VideoMAE                  & Video & 0.880 & 0.820 & 0.728 & 0.748 & 0.962 & 0.647 & 0.634 \\
        \bottomrule
    \end{tabular}
    \caption{Blink detection performance for binary (full-blink + half-blink vs.\ none) and three-class (none / half-blink / full-blink) tasks. Per-class F1 is not reported for binary classification (-).}
    \label{tab:blink-results}
\end{table*}

\begin{table*}[htbp]
    \footnotesize
    \centering
    \begin{tabular}{lcccccccc}
    \toprule
     & & \multicolumn{2}{c}{\% Agreement} & & \multicolumn{2}{c}{start diff (frames)} & \multicolumn{2}{c}{end diff (frames)} \\
    \cmidrule(lr){3-4} \cmidrule(lr){6-7} \cmidrule(lr){8-9}
    Class & $N$ & $\mu$ & $\sigma$ & Krippendorff's $\alpha$ & $\mu$ & $\sigma$ & $\mu$ & $\sigma$ \\
    \midrule
    None    &  8 & .94 & .12 & .90              & ---  & ---  & ---  & ---  \\
    Half-blink &  5 & .73 & .17 & .57              & 2.07 & 3.77 & 1.96 & 1.59 \\
    Blink      &  9 & .94 & .08 & .73              & 0.95 & 0.84 & 1.92 & 1.94 \\
    \midrule
    Overall    & 22 & .89 & .15 & .85              & 1.35 & 2.41 & 1.93 & 1.82 \\
    \bottomrule
    \end{tabular}
    \caption{Inter-rater reliability by class: categorical agreement and temporal boundary consistency. $N$ represents the number of clips each rater classified. Mean is represented by $\mu$ and standard deviation by $\sigma$.}
    \label{tab:inter_study1}
\end{table*}

\pgfplotsset{compat=1.18}
\begin{figure}[htbp]
    \centering
    \begin{tikzpicture}
    \begin{axis}[
        ybar,
        bar width=0.43cm,
        width=0.5\textwidth,
        height=6cm,
        ymin=0, ymax=1.1,
        ytick={0, 0.2, 0.4, 0.6, 0.8, 1.0},
        yticklabel={\pgfmathprintnumber[fixed, precision=1]{\tick}},
        ylabel={F1 Score},
        symbolic x coords={None, Half-blink, Full-blink, Macro-F1},
        xtick=data,
        xticklabel style={font=\small},
        legend style={
            at={(0.5,1.02)},
            anchor=south,
            legend columns=2,
            font=\small,
        },
        enlarge x limits=0.2,
        nodes near coords,
        nodes near coords style={font=\scriptsize},
        every node near coord/.append ,
    ]
    
    \addplot[fill=gray!40, draw=black] coordinates {
        (None,    0.81)
        (Half-blink, 0.59)
        (Full-blink, 0.89)
        (Macro-F1,    0.76)
    };
    
    \addplot[fill=black!75, draw=black] coordinates {
        (None,    1.00)
        (Half-blink, 1.00)
        (Full-blink, 1.00)
        (Macro-F1,    1.00)
    };
    
    \legend{Human annotators, YOLO}
    
    \end{axis}
    \end{tikzpicture}
    \caption{F1 score comparison naive human annotators and the YOLO model across blink classes on the inter-annotator clip set (22 clips).}
    \label{fig:human_vs_yolo_f1}
\end{figure}
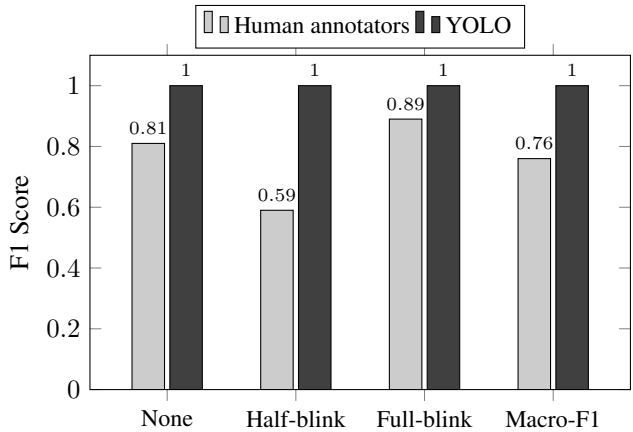

\begin{figure*}
    \centering
    \includegraphics[width=\linewidth]{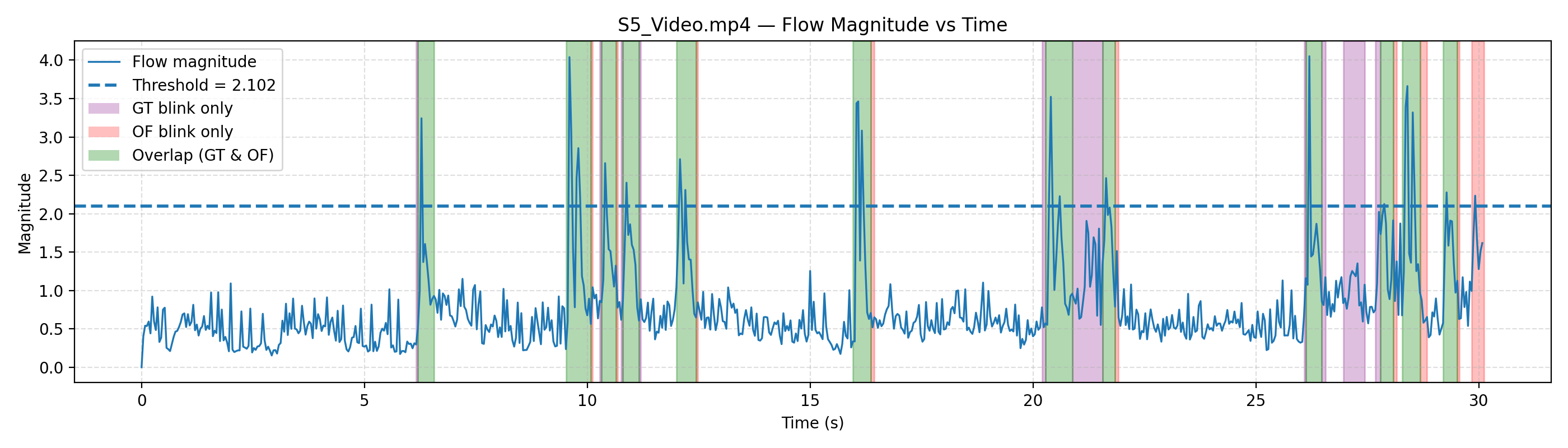}
    \caption{Optical flow based method on example full-length S5 video showing how flow magnitude contributes for blink detection.}
    \label{fig:quant_mag_opticflow}
\end{figure*}

\begin{figure}
    \centering
    \begin{subfigure}{\linewidth}
        \centering
        \includegraphics[width=\textwidth]{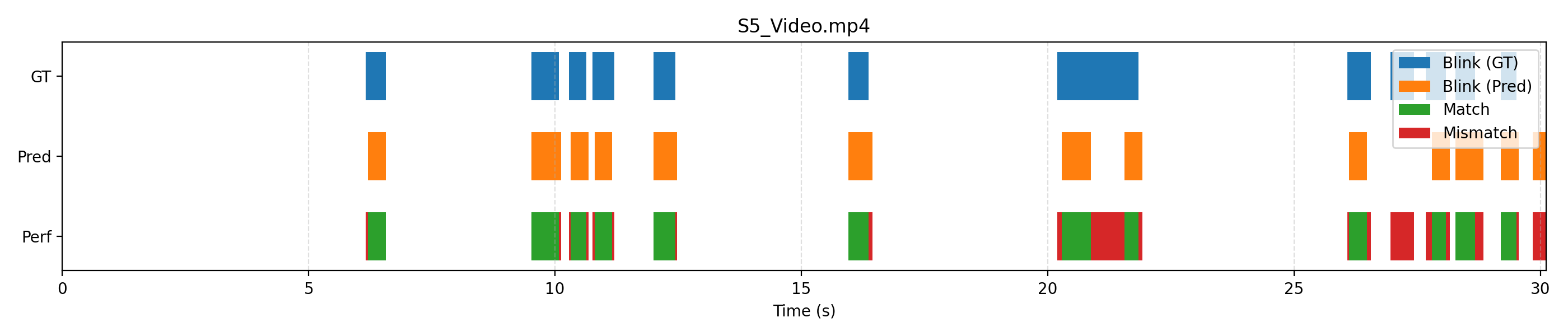}
        \caption*{(a)}
        \label{fig:qual_opticflow}
    \end{subfigure}
    \begin{subfigure}{\linewidth}
        \centering
        \includegraphics[width=\textwidth]{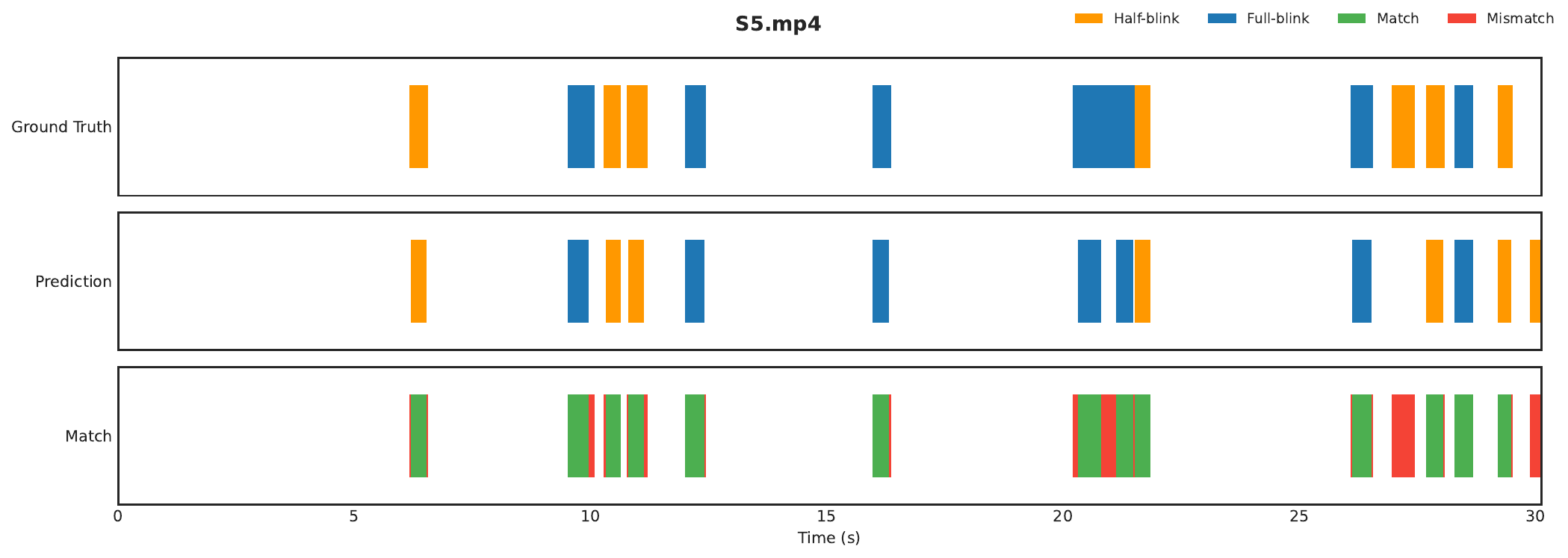}
        \caption*{(b)}
        \label{fig:qual_yolo}
    \end{subfigure}
    \begin{subfigure}{\linewidth}
        \centering
        \includegraphics[width=\textwidth]{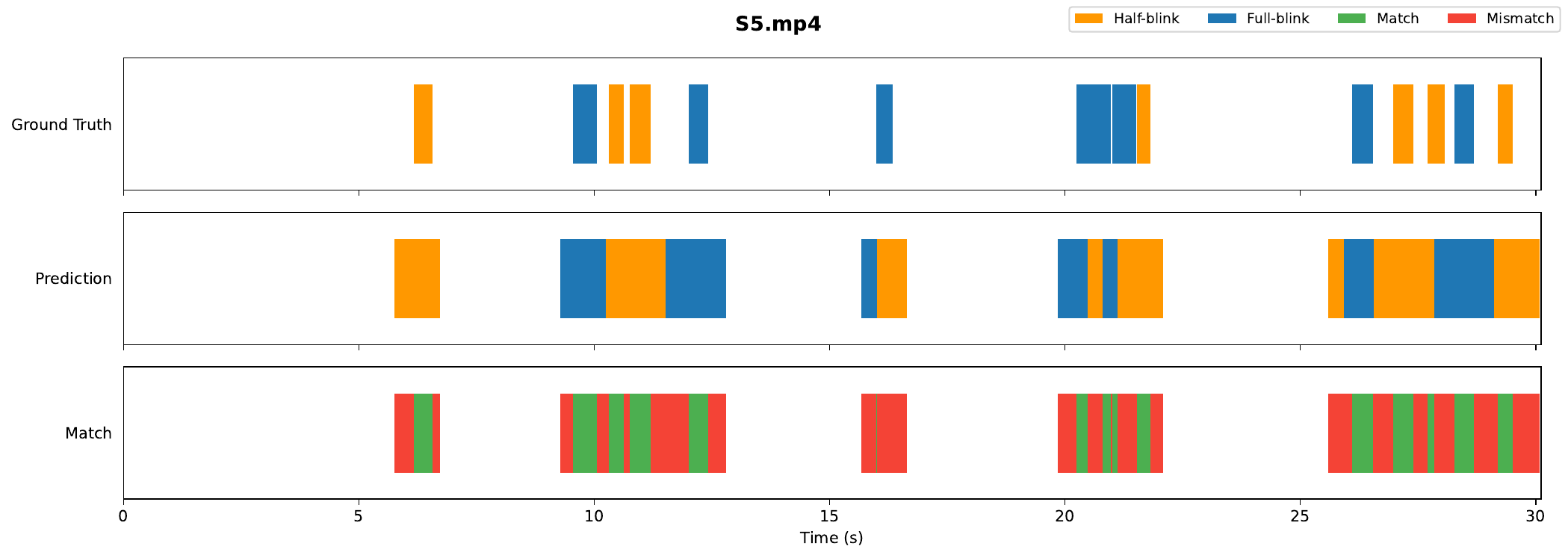}
        \caption*{(c)}
        \label{fig:qual_videomae}
    \end{subfigure}
    \caption{Example qualitative results on original S5 video for (a) optical flow method, (b) YOLO based method and (c) VideoMAE method. Additional results in supplementary material.}
    \label{fig:quali_results}
\end{figure}

\section{Experiments and Results} \label{experiments}

\subsection{Inter-annotator study}\label{interannotatorstudy}
To establish a human baseline, we conducted an annotation study on data from the expert annotated public test set~\cite{rashidEquineFacialAction2020} via a web-based interface where naive participants, provided with class descriptions, classified 22 short horse video clips (selected from the test set to avoid ambiguity; see Fig. \ref{fig:blink_dist}) as \textit{full-blink}, \textit{half-blink}, or \textit{none}, and later annotated start and end frames for blink classes. A total of $N=6$ participants completed the study. Overall classification accuracy was 78.8\% (macro-F1=0.76, Krippendorff's $\alpha$=0.85). Agreement was high for \textit{full-blink} (94\%, $\alpha$=0.73) and \textit{none} (94\%, $\alpha$=0.90), but substantially lower for \textit{half-blink} (73\%, $\alpha$=0.57), with 10 of 30 half-blink annotations misclassified as full-blink, reflecting the ambiguity of partial eyelid closure. Temporal annotations show a mean absolute error of 1.75 frames for start and 2.88 frames for end boundaries. Additional data can be found in Table~\ref{tab:inter_study1} and \ref{tab:inter_study2} in the supplementary material.

\subsection{Quantitative results}
Regarding binary blink detection, both VideoMAE and YOLO achieve a macro-F1 score of 0.926 on the video-clip-based test set, while YOLO attains 0.870 on the frame-based test set. The optical flow method shows misleadingly high accuracy due to class imbalance in the dataset, reflected in its macro-F1 score of 0.680. For the three-class classification task, YOLO achieves macro-F1 scores of 0.782 and 0.898 on the frame-based and video-based test sets, respectively, while VideoMAE achieves 0.748 on the video-based set. These results demonstrate the complexity of blink-type classification in horses, consistent with our human baseline. Detailed results are presented in Table~\ref{tab:blink-results} and Figures~\ref{fig:CM_det} and \ref{fig:CM_class}. We evaluate our best performing model (YOLO) on the inter-annotator study's 22 video clips for fair comparison with our naive human annotators. Our model achieves perfect F1 scores across all classes on the inter-annotator study's 22 video clips (partly explained by the low-ambiguity selection of the clips) which validates the method's efficacy (see Figure \ref{fig:human_vs_yolo_f1}).

\subsection{Qualitative results}
To further evaluate performance in a real world application, we naively classify our methods on the original full-length horse videos using the three methods. Figures \ref{fig:quali_results} and fig\ref{fig:quant_mag_opticflow} qualitatively show how the different methods fared on an example test video~\cite{gleerupEquinePainFace2015}. We see that the VideoMAE based method tends to have a higher recall of blink events. This is related to the window sliding approach taken with Video MAE on the original video data, producing less granular predictions when compared with the frame based methods. However, in semi-automated annotation, a higher false positive rate is often preferable to an excess of false negatives, as the latter could compromise completeness, as long as it does not hinder feasibility and speed.

%% file: sections/5Discussion_Conclusion.tex
\section{Discussion \& Conclusion}

Our results show that equine blink detection can be effectively automated across multiple approaches. On binary detection, VideoMAE and YOLO achieve a macro-F1 of 0.926 with the video test set, while YOLO attains 0.870 and optical flow 0.680 with the frame-based test set. On the three-class problem, YOLO achieves 0.898, exceeding the human annotator baseline of 0.760, while VideoMAE reaches 0.748. The high recall of learned models aligns with the false positive preference mentioned previously, ensuring few blinks are missed while offering a scalable alternative to manual labelling. Each approach carries limitations. VideoMAE requires a fixed input of 16 frames, reducing temporal resolution in longer sequences, while the YOLO-based method discards temporal context by operating on individual frames, limiting sensitivity to subtle motion. The optical flow approach is sensitive to hyperparameter choices, reducing robustness across varied conditions. Across all methods, half-blink classification remains the central challenge, mirroring low human inter-rater agreement. This reflects the perceptual ambiguity of partial eyelid closure, at least when observing video at normal speed. In summary, we present the first systematic comparison of machine learning methods for equine blink detection, supported by a human baseline study. Future work should pursue larger datasets and broader pain-relevant facial action units.

\paragraph{Code availability.} The code and supplementary materials used in this research work will be made available at \url{https://github.com/jmalves5/read-my-eyes}.

\paragraph{Acknowledgements.} This work has been funded by the Independent Research Fund Denmark under grant ID 10.46540/3105-00114B.

%% file: sections/suppl.tex
\clearpage
\setcounter{page}{1}
\maketitlesupplementary

This document contains the supplementary material for the CVPR 2026 CV4Animals workshop paper titled \textit{``Horse Eye Blink Detection and Classification for Equine Affective State Assessment''} and provides further insight into the results obtained with the different methods tested.

\section{Dataset subjects sample data}
\begin{figure*}[htbp]
    \centering
    \begin{subfigure}{0.32\linewidth}
        \includegraphics[width=\linewidth]{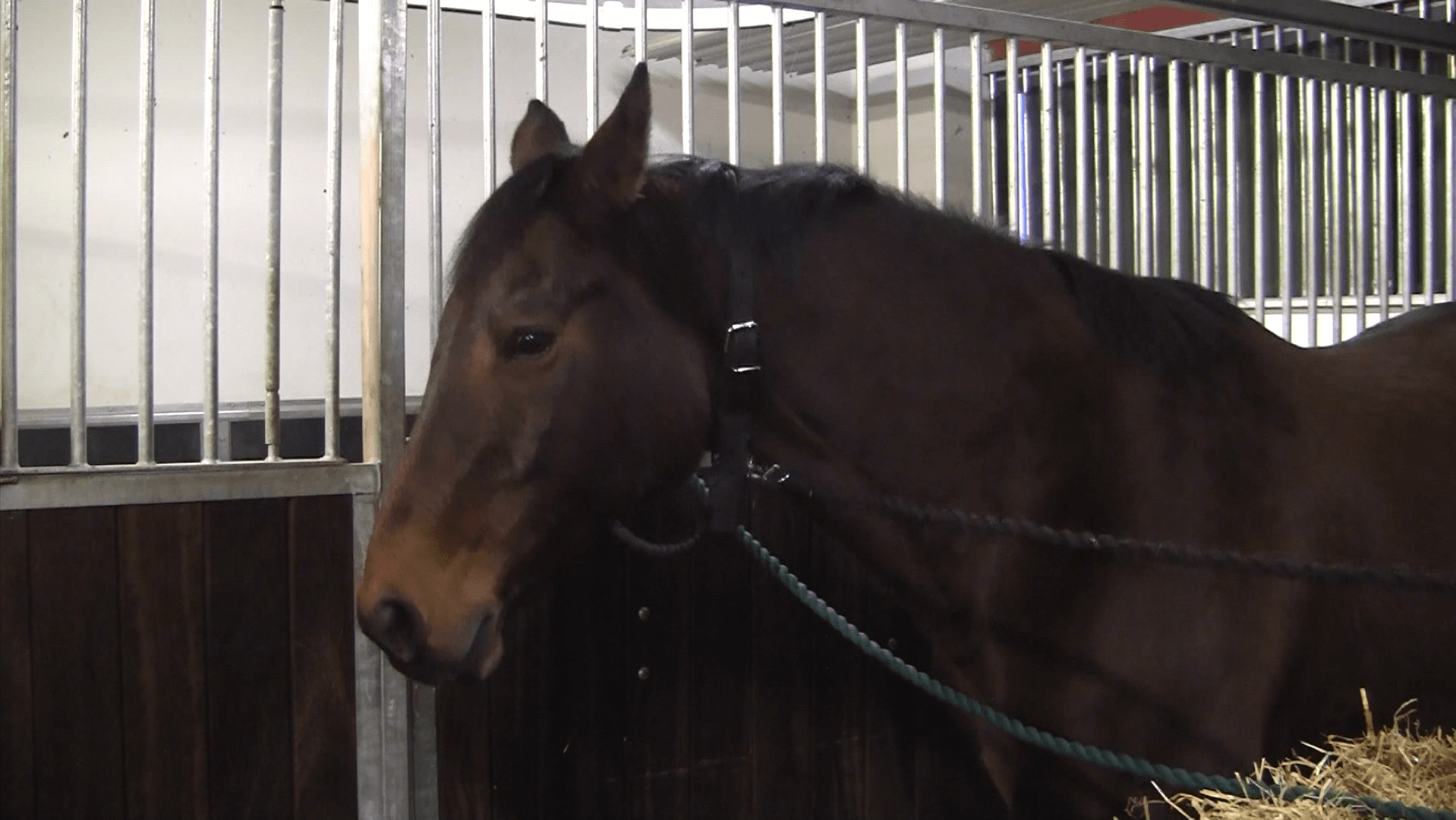}
    \end{subfigure}
    \begin{subfigure}{0.32\linewidth}
        \includegraphics[width=\linewidth]{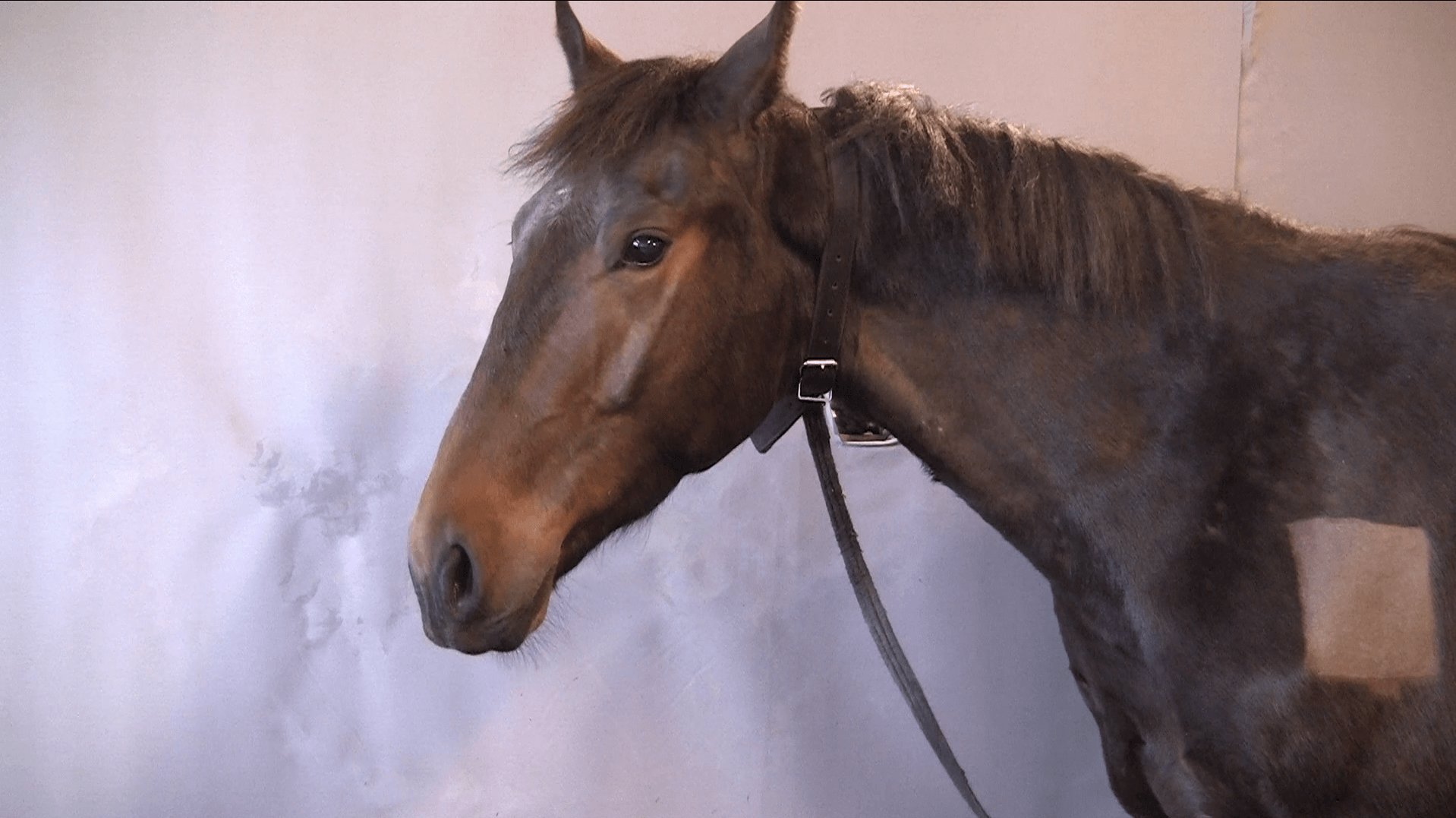}
    \end{subfigure}
    \begin{subfigure}{0.32\linewidth}
        \includegraphics[width=\linewidth]{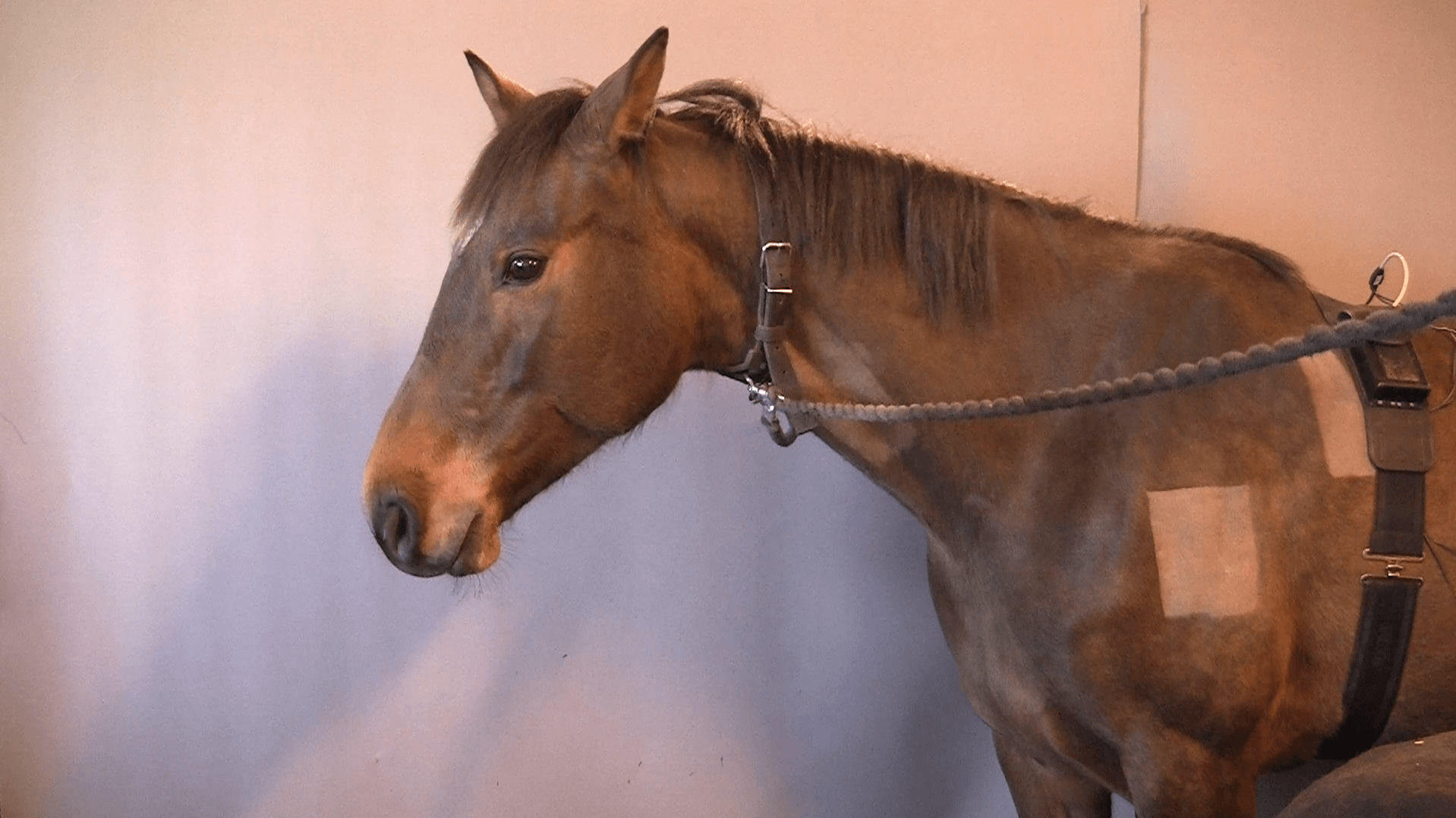}
    \end{subfigure}
    \\[0.5em]

    \begin{subfigure}{0.32\linewidth}
        \includegraphics[width=\linewidth]{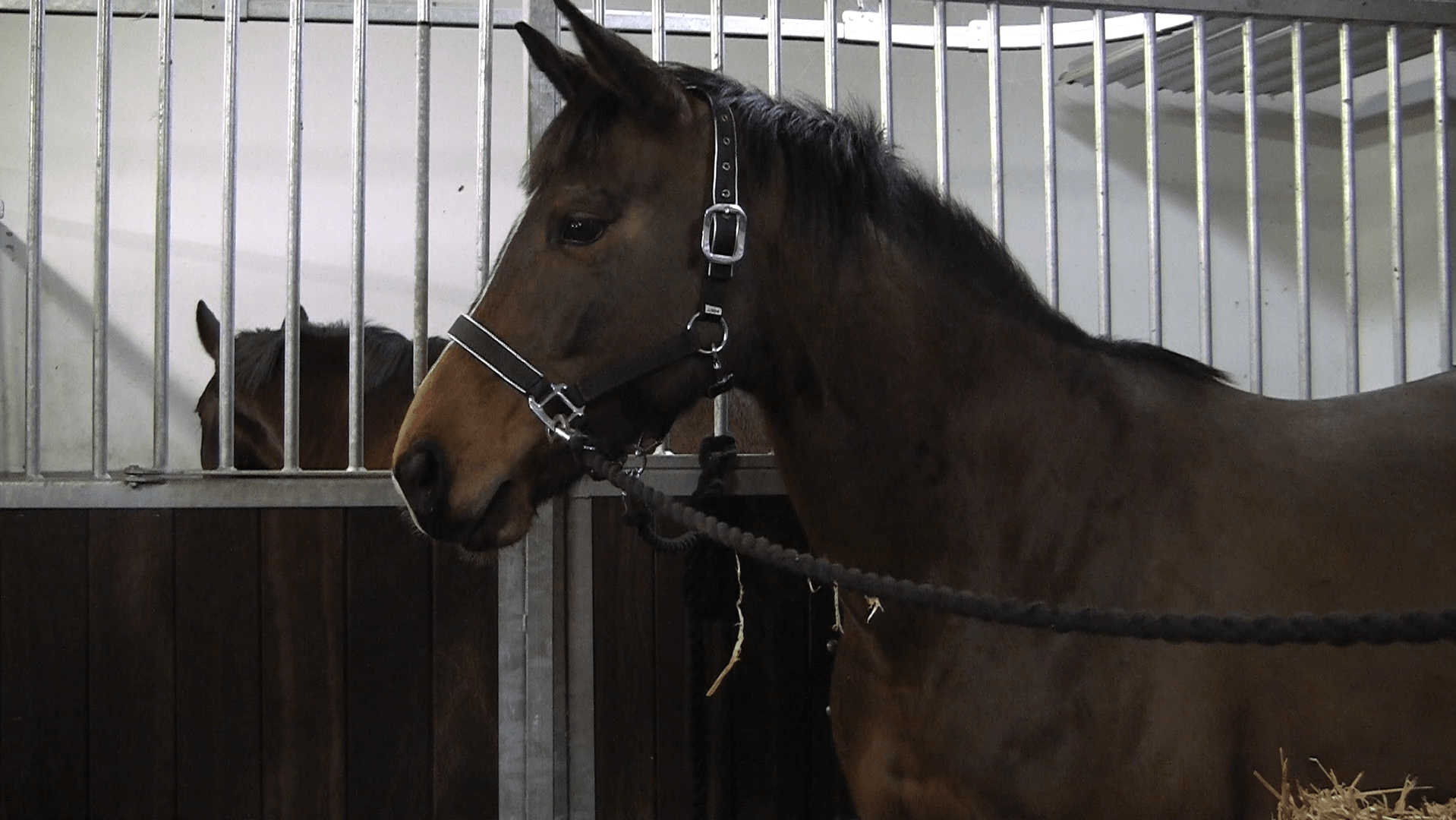}
    \end{subfigure}
    \begin{subfigure}{0.32\linewidth}
        \includegraphics[width=\linewidth]{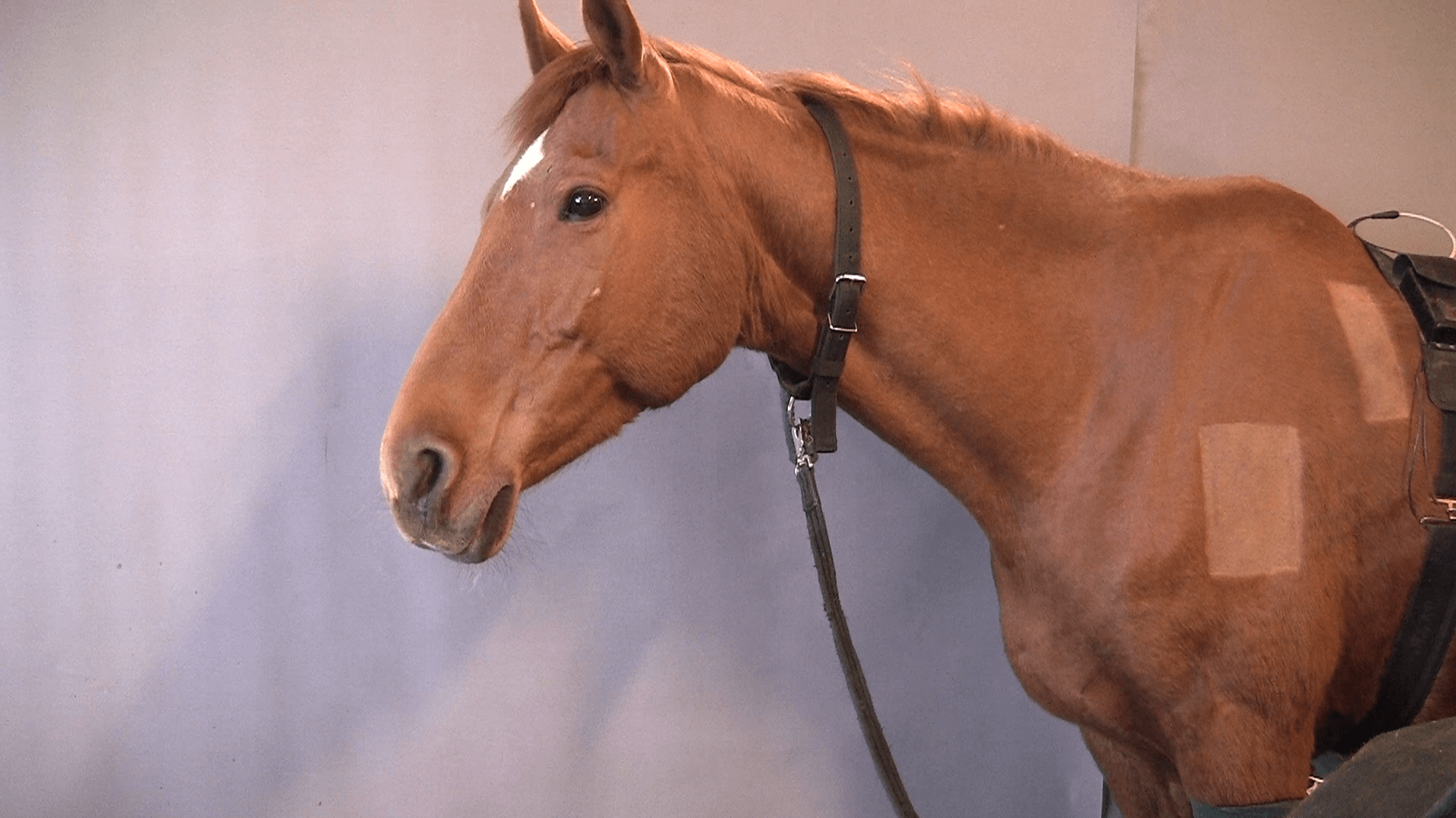}
    \end{subfigure}
    \begin{subfigure}{0.32\linewidth}
        \includegraphics[width=\linewidth]{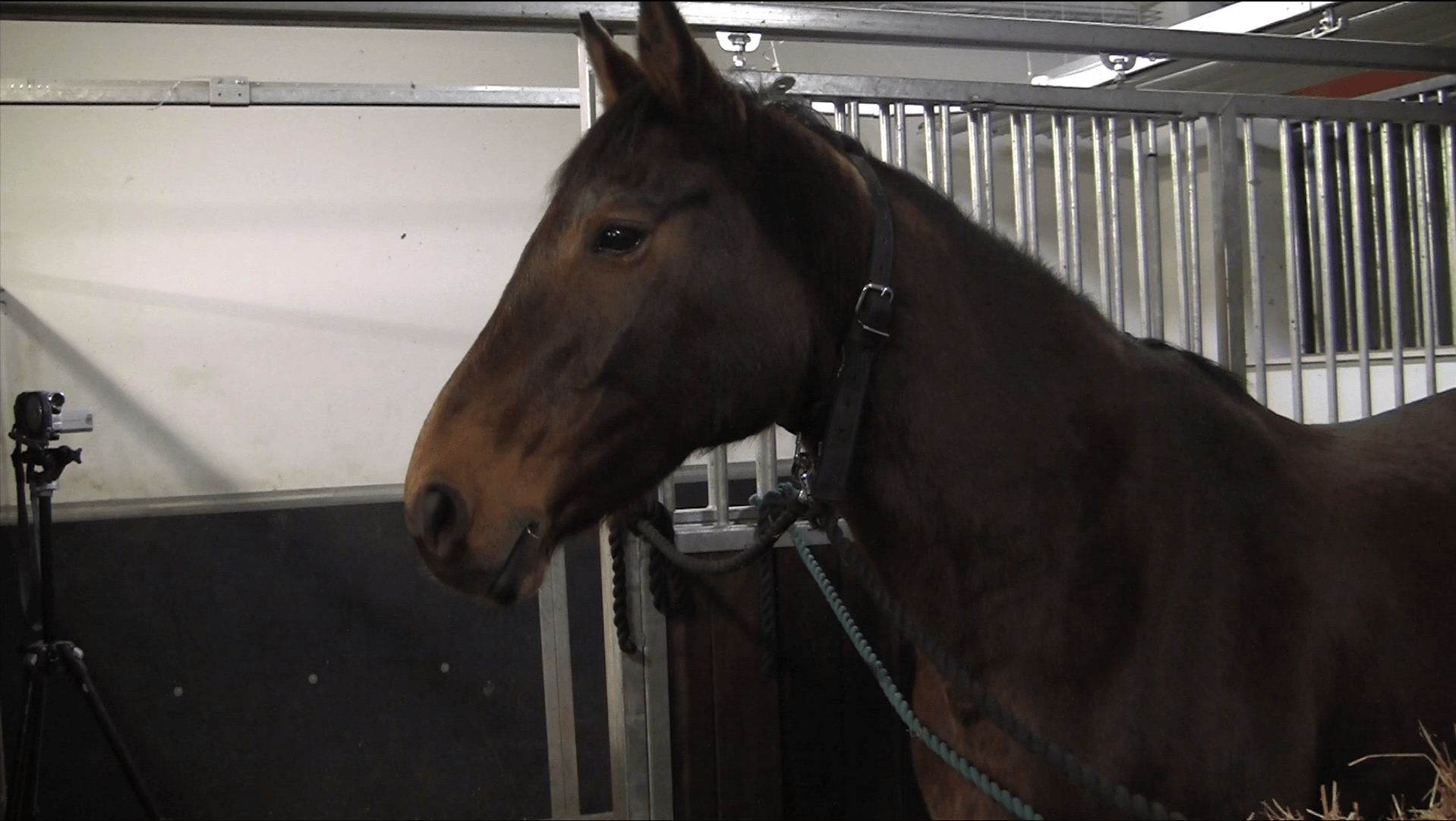}
    \end{subfigure}
    \\[0.5em]

    \begin{subfigure}{0.32\linewidth}
        \includegraphics[width=\linewidth]{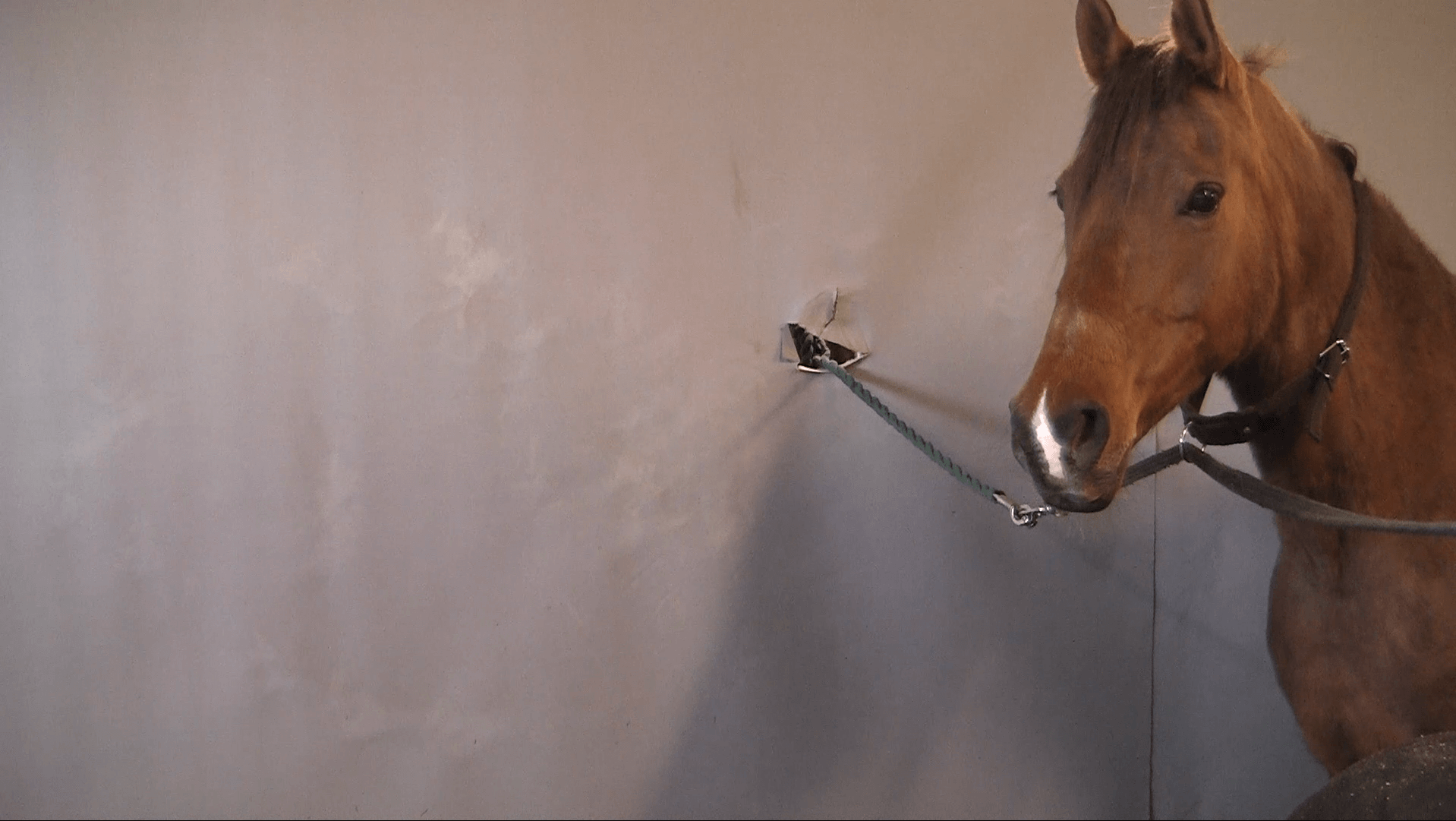}
    \end{subfigure}
    \begin{subfigure}{0.32\linewidth}
        \includegraphics[width=\linewidth]{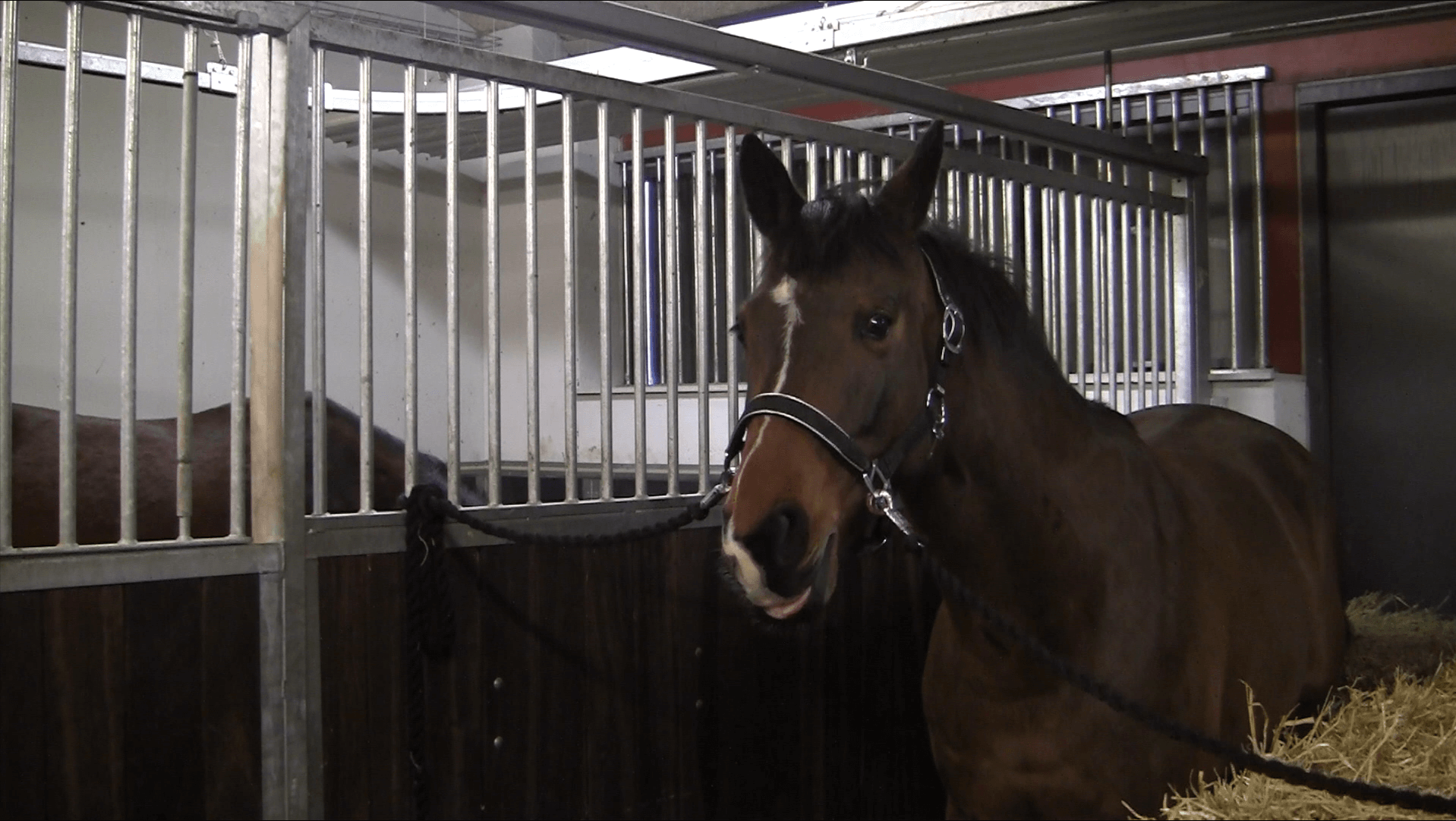}
    \end{subfigure}
    \begin{subfigure}{0.32\linewidth}
        \includegraphics[width=\linewidth]{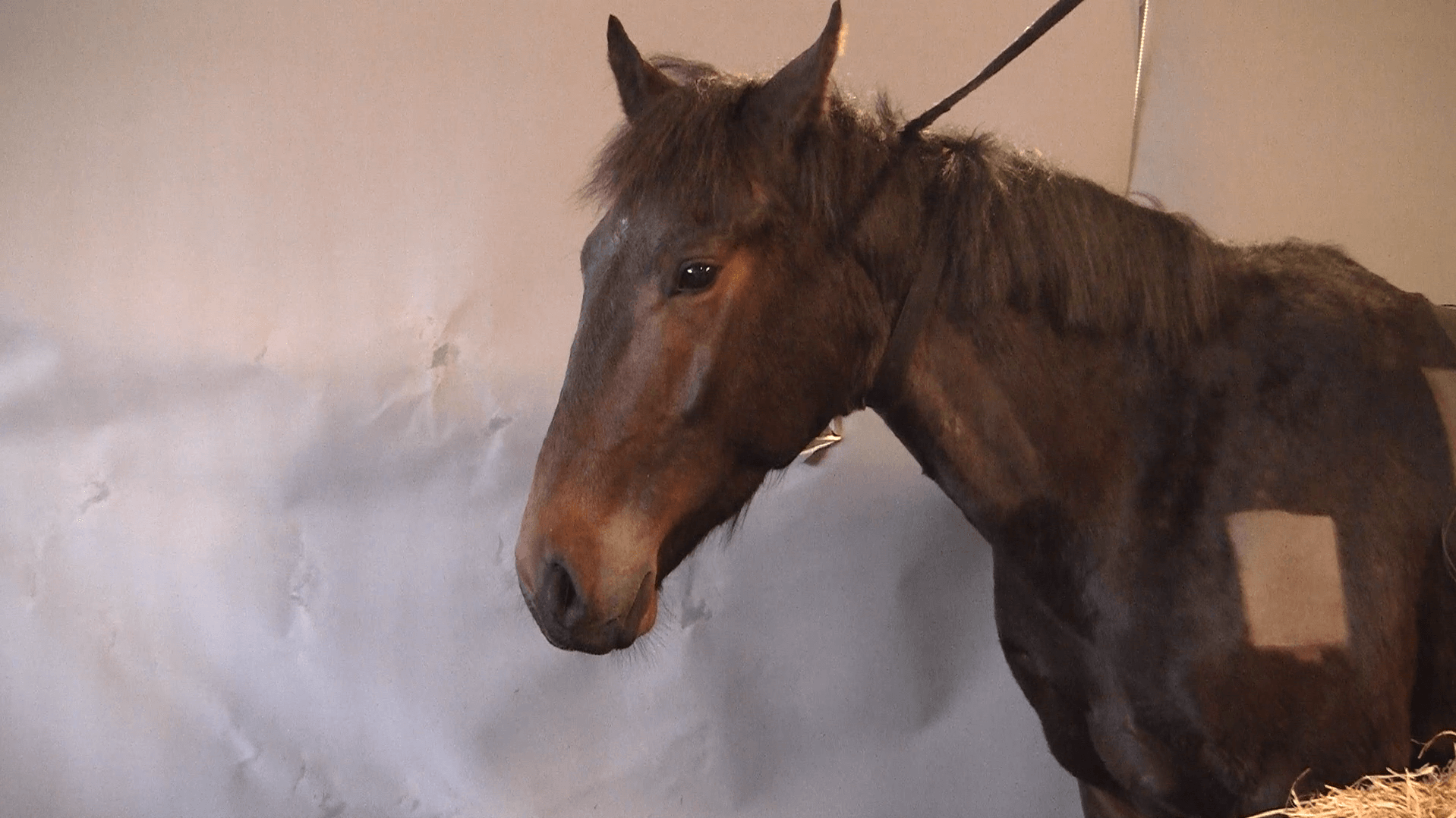}
    \end{subfigure}
    \\[0.5em]

    \begin{subfigure}{0.32\linewidth}
        \includegraphics[width=\linewidth]{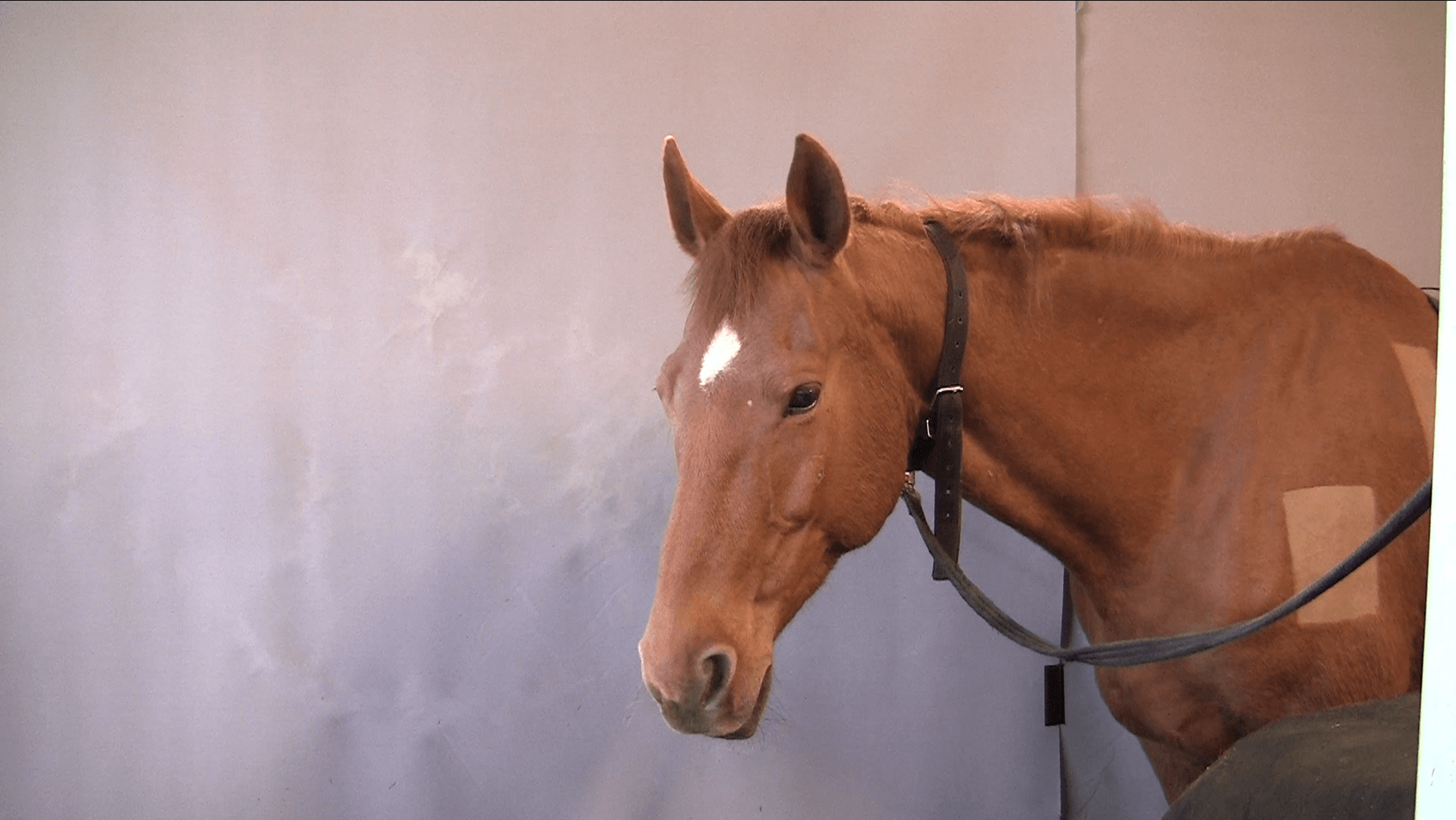}
    \end{subfigure}
    \begin{subfigure}{0.32\linewidth}
        \includegraphics[width=\linewidth]{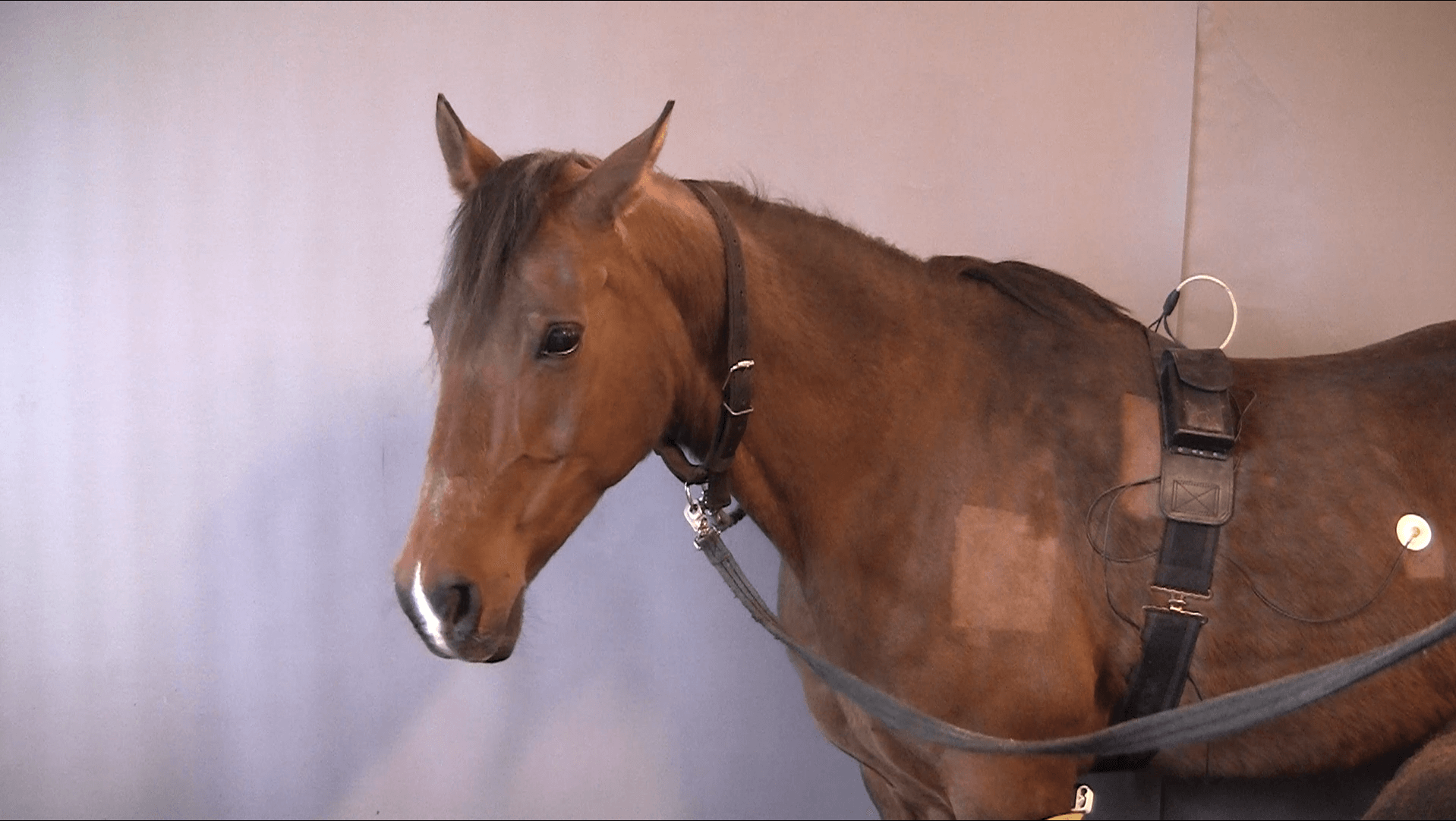}
    \end{subfigure}
    \begin{subfigure}{0.32\linewidth}
        \includegraphics[width=\linewidth]{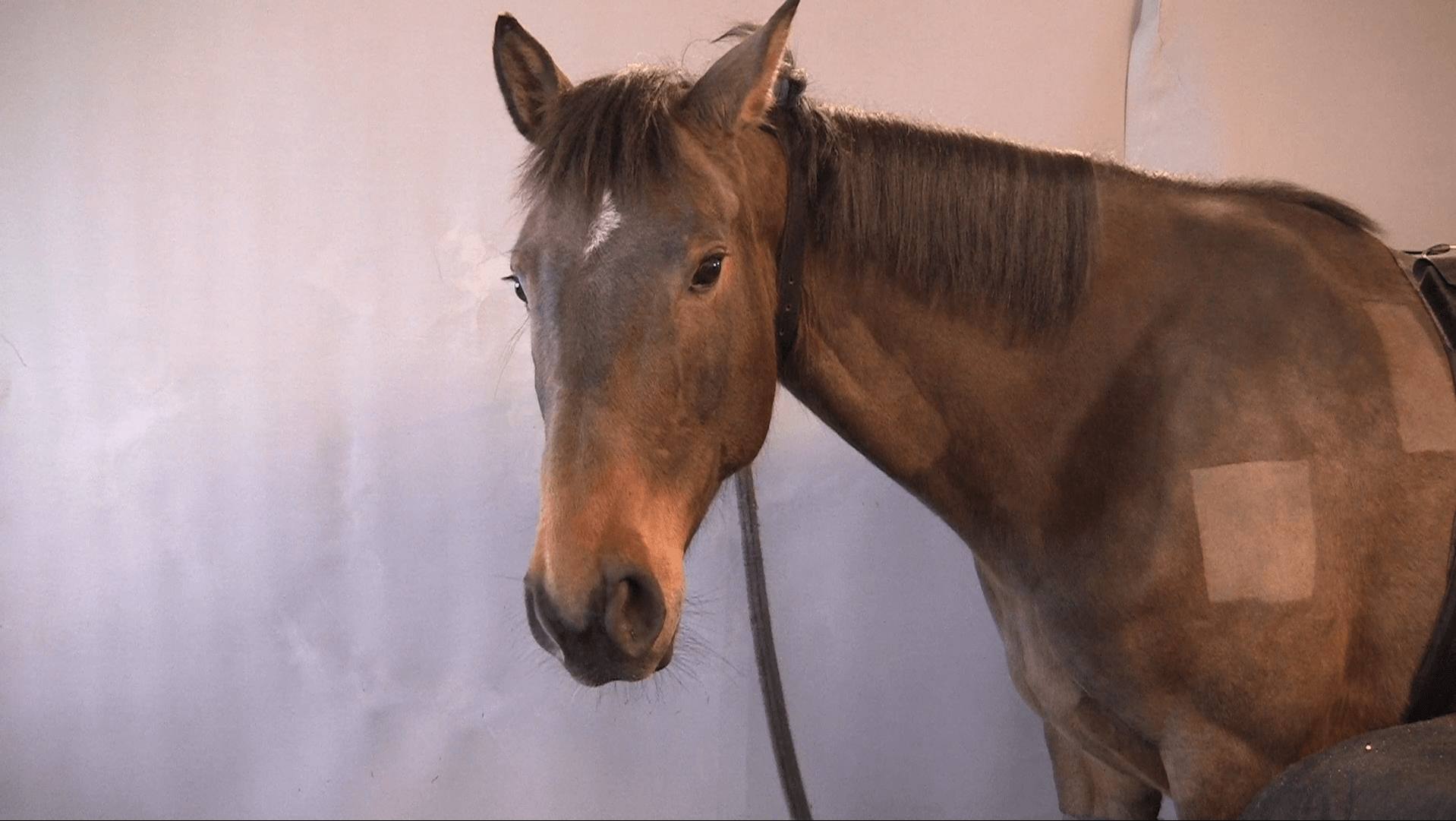}
    \end{subfigure}
    \caption{Sample frames for each of the 12 videos \cite{gleerupEquinePainFace2015} in the test dataset in row-major order.}
    \label{fig:sample_frames}
\end{figure*}

To provide further insight into the dataset used, we provide sample frames of each of the 12 test videos from~\cite{rashidEquineFacialAction2020} in this supplementary material (see Figure \ref{fig:sample_frames}).

\section{Human baseline study}
Additional results of the human baseline study are presented in Table \ref{tab:inter_study2}.

\section{Supplementary qualitative results}
\begin{figure*}[htbp]
    \centering
    \begin{subfigure}{0.49\linewidth}
        \includegraphics[width=\linewidth]{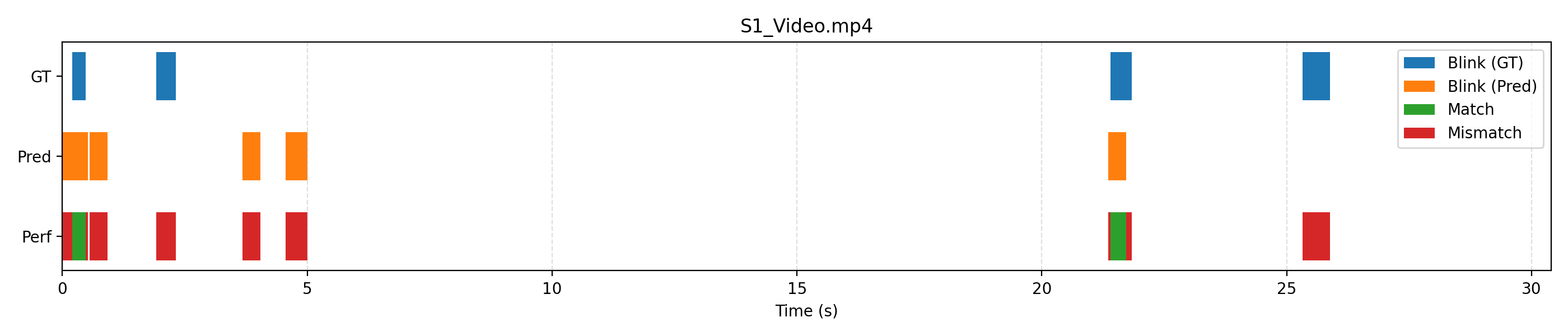}
    \end{subfigure}
    \begin{subfigure}{0.49\linewidth}
        \includegraphics[width=\linewidth]{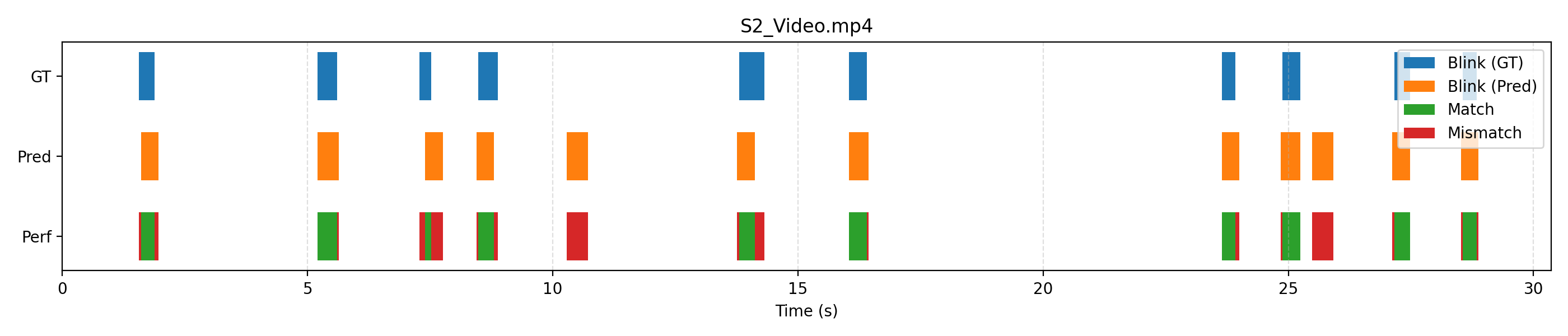}
    \end{subfigure}
    \\[0.5em]
    \begin{subfigure}{0.49\linewidth}
        \includegraphics[width=\linewidth]{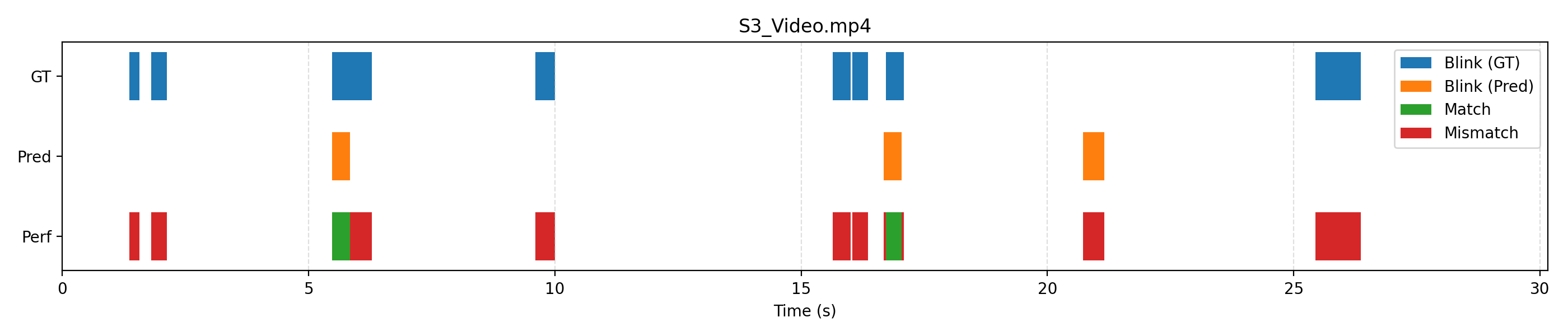}
    \end{subfigure}
    \begin{subfigure}{0.49\linewidth}
        \includegraphics[width=\linewidth]{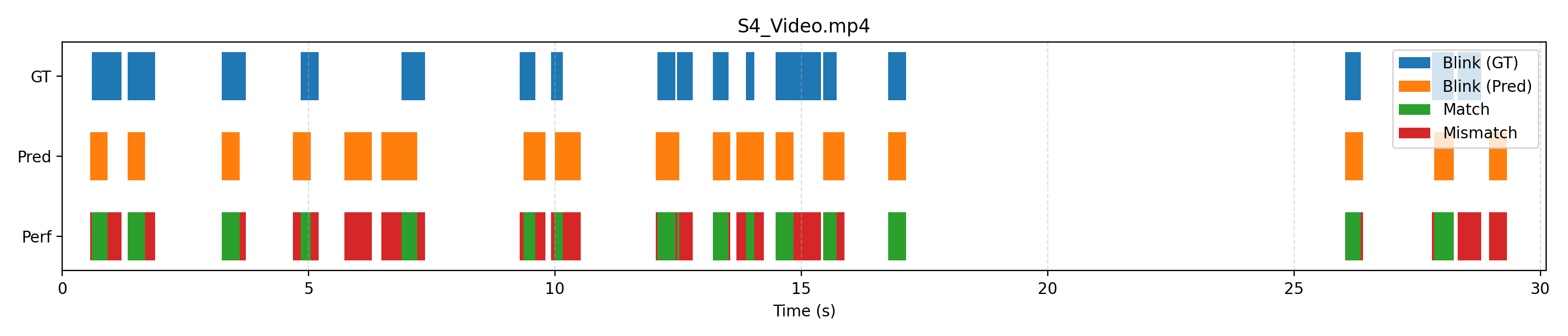}
    \end{subfigure}
    \\[0.5em]
    \begin{subfigure}{0.49\linewidth}
        \includegraphics[width=\linewidth]{images/OpticFlow/timeline/S5_Video_timeline_GT_Pred_Perf.png}
    \end{subfigure}
    \begin{subfigure}{0.49\linewidth}
        \includegraphics[width=\linewidth]{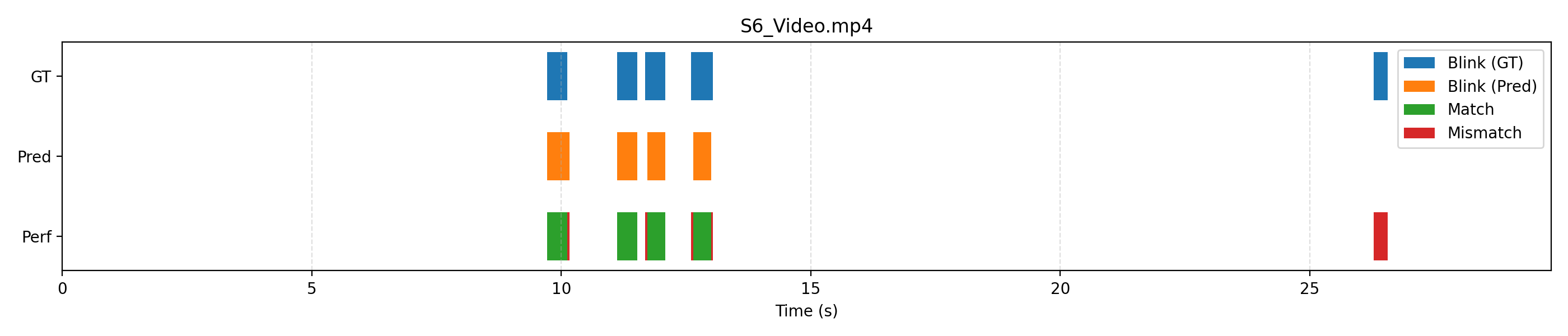}
    \end{subfigure}
    \\[0.5em]
    \begin{subfigure}{0.49\linewidth}
        \includegraphics[width=\linewidth]{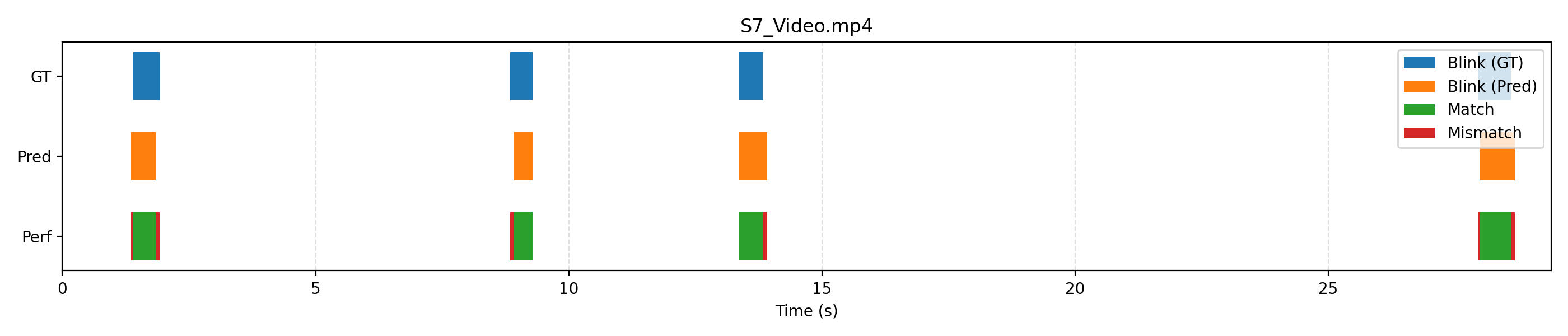}
    \end{subfigure}
    \begin{subfigure}{0.49\linewidth}
        \includegraphics[width=\linewidth]{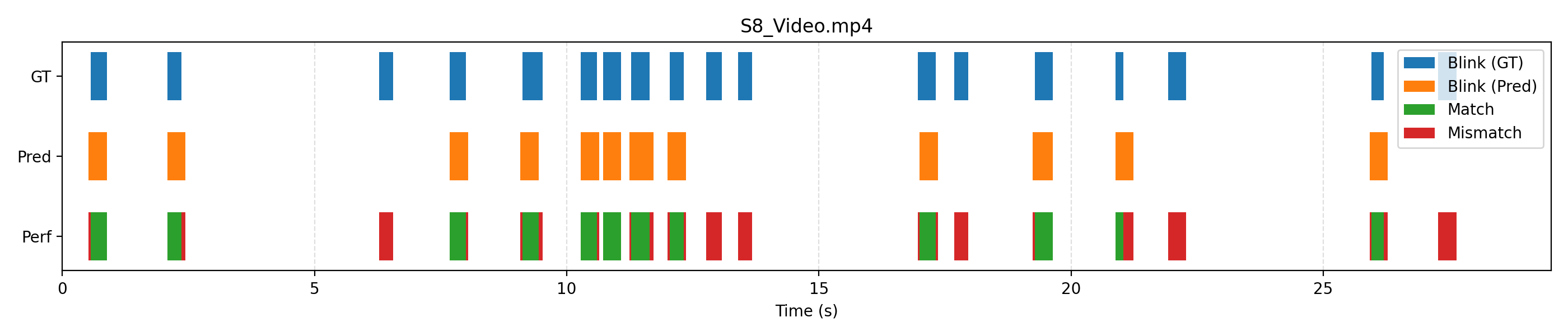}
    \end{subfigure}
    \\[0.5em]
    \begin{subfigure}{0.49\linewidth}
        \includegraphics[width=\linewidth]{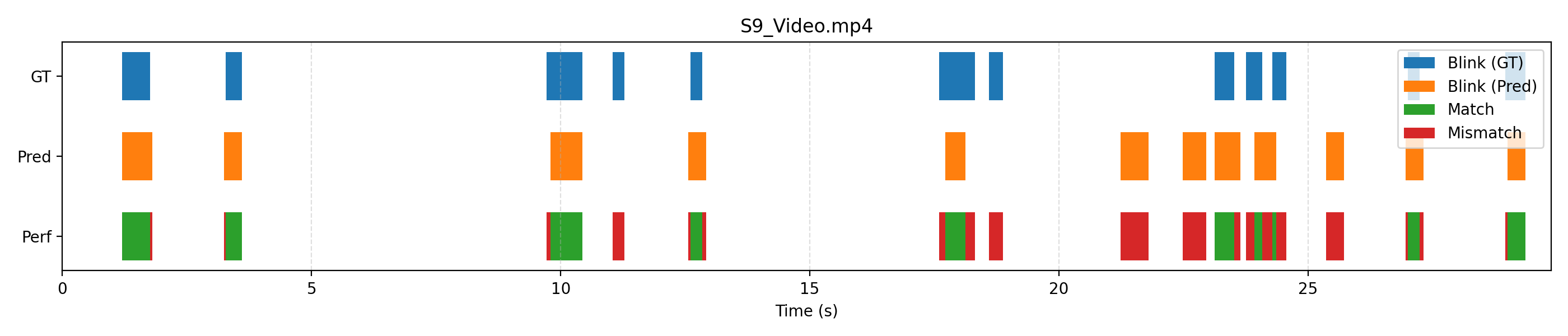}
    \end{subfigure}
    \begin{subfigure}{0.49\linewidth}
        \includegraphics[width=\linewidth]{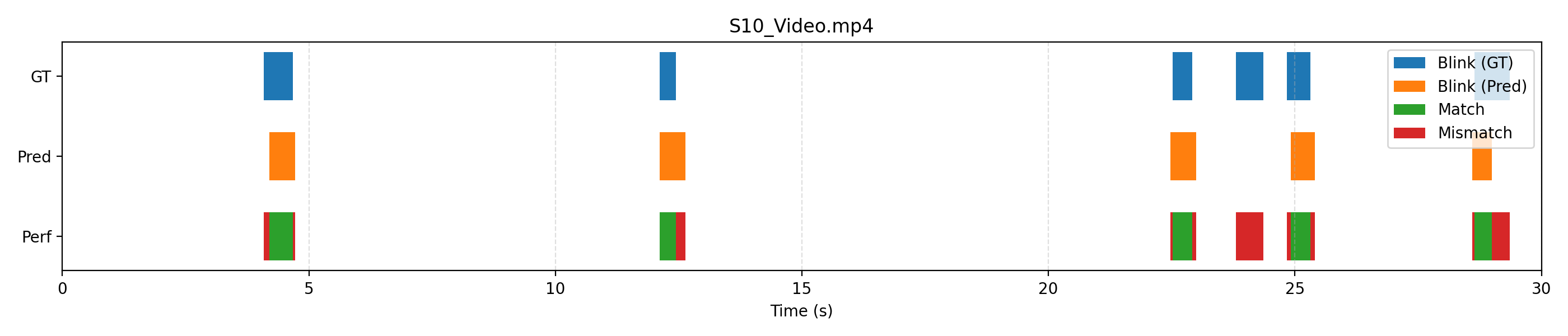}
    \end{subfigure}
    \\[0.5em]
    \begin{subfigure}{0.49\linewidth}
        \includegraphics[width=\linewidth]{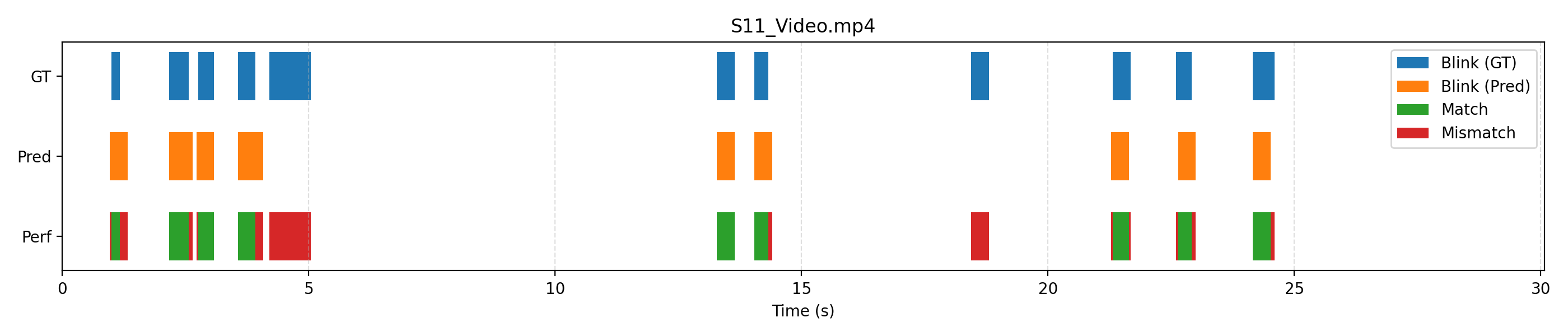}
    \end{subfigure}
    \begin{subfigure}{0.49\linewidth}
        \includegraphics[width=\linewidth]{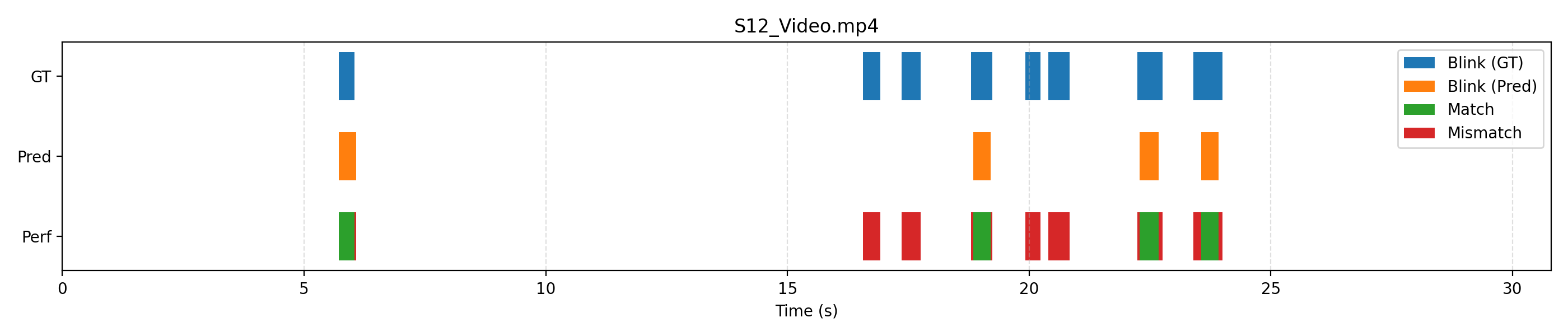}
    \end{subfigure}
    \caption{Qualitative video timelines (S1-S12) for Optical Flow based eyelid closure detection method.}
    \label{fig:of_quali}
\end{figure*}

\begin{figure*}[htbp]
    \centering
    \begin{subfigure}{0.49\linewidth}
        \includegraphics[width=\linewidth]{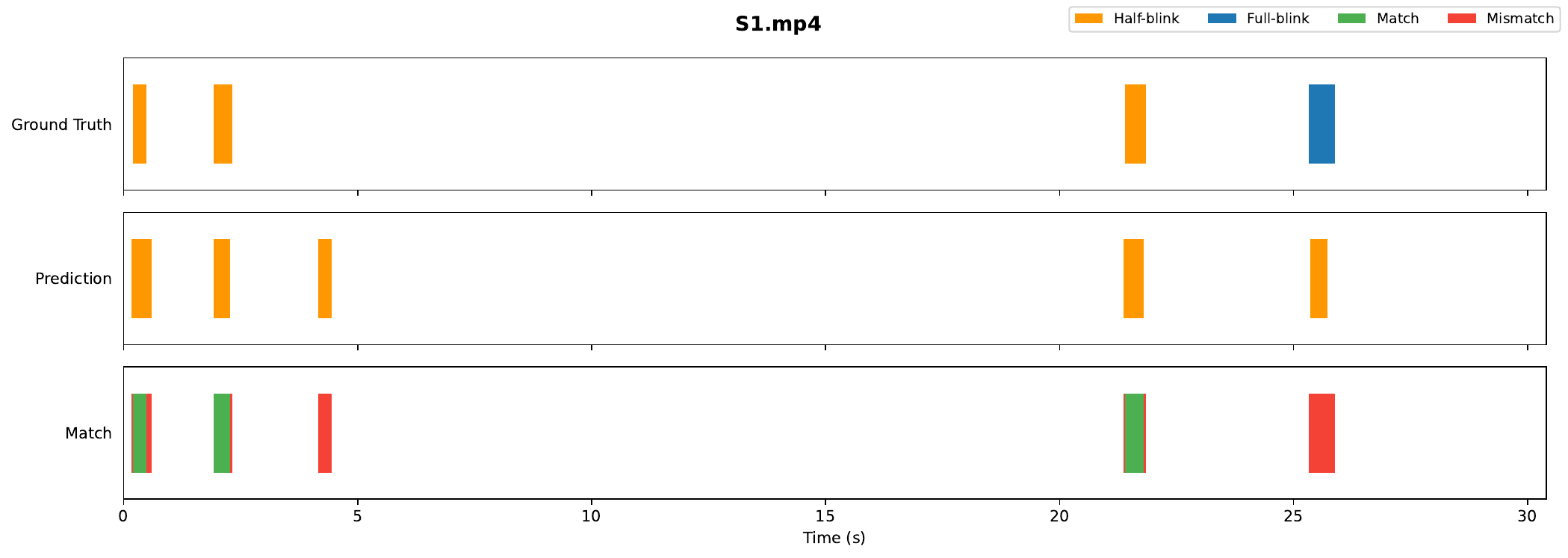}
    \end{subfigure}
    \begin{subfigure}{0.49\linewidth}
        \includegraphics[width=\linewidth]{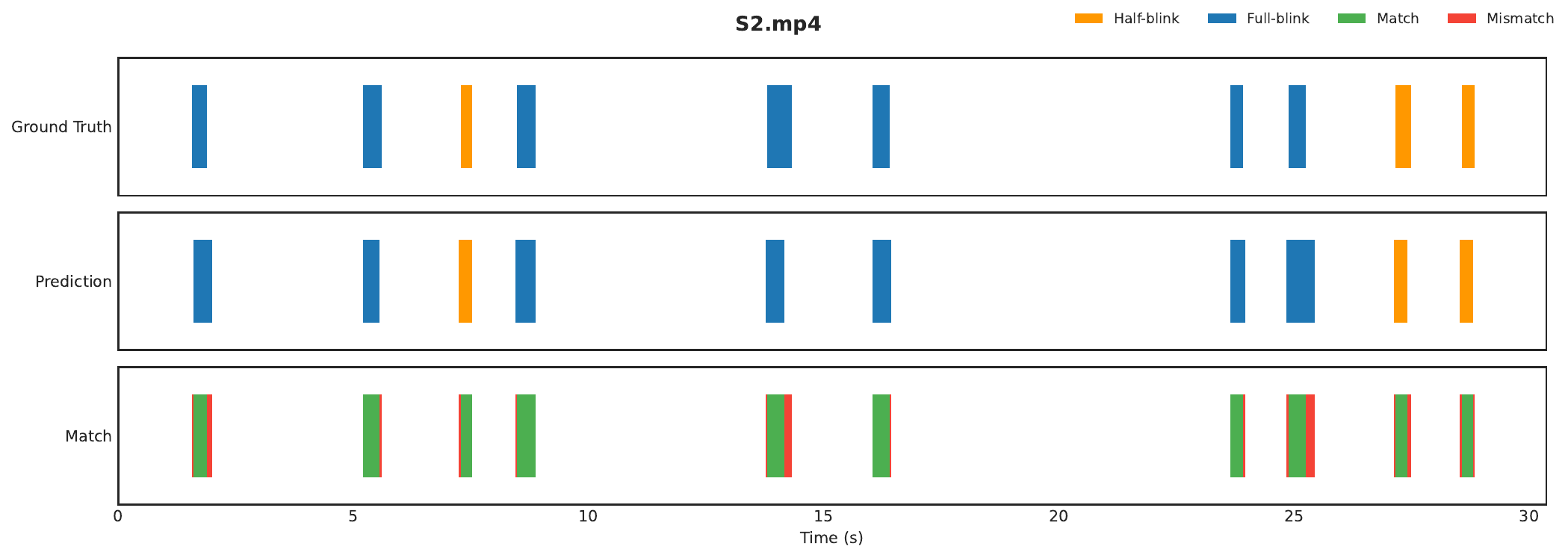}
    \end{subfigure}
    \\[0.5em]
    \begin{subfigure}{0.49\linewidth}
        \includegraphics[width=\linewidth]{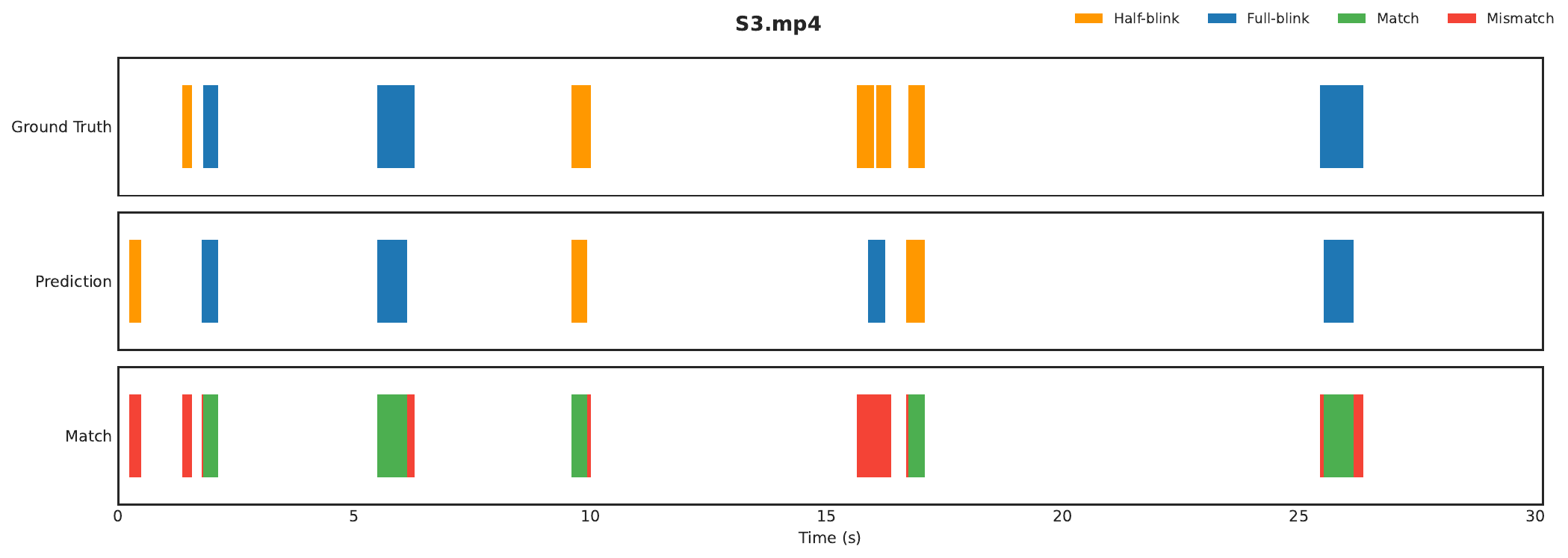}
    \end{subfigure}
    \begin{subfigure}{0.49\linewidth}
        \includegraphics[width=\linewidth]{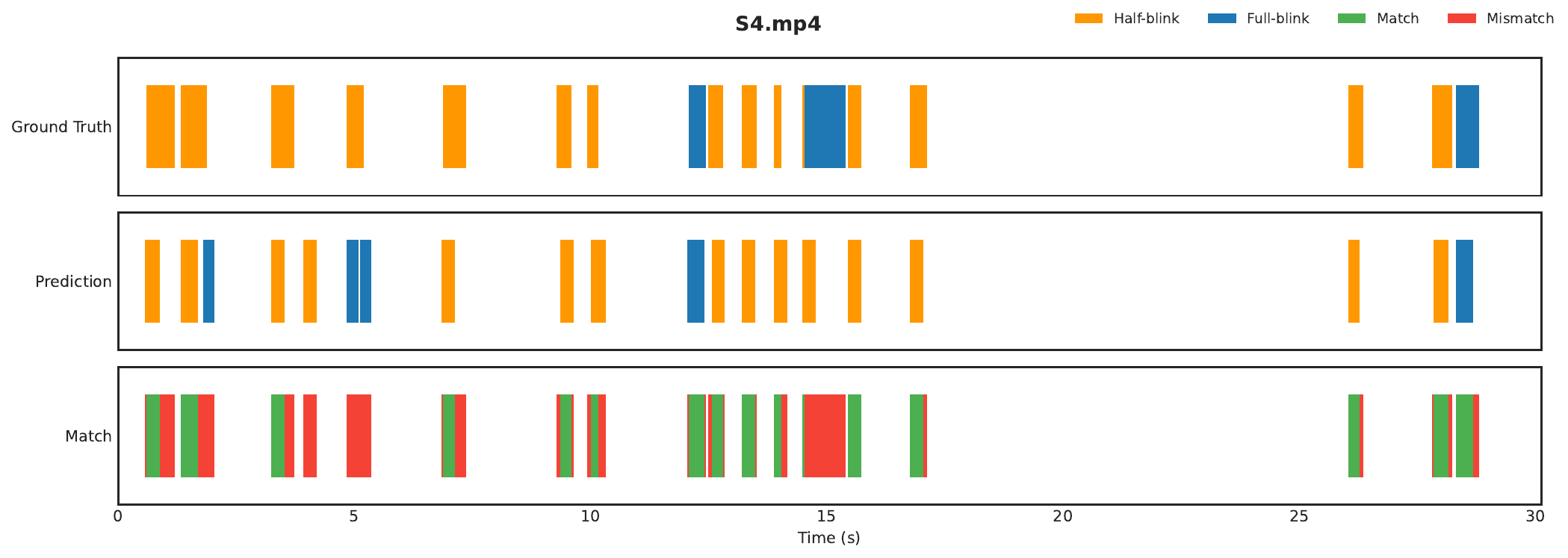}
    \end{subfigure}
    \\[0.5em]
    \begin{subfigure}{0.49\linewidth}
        \includegraphics[width=\linewidth]{images/YOLO/timeline/S5_Video_timeline.pdf}
    \end{subfigure}
    \begin{subfigure}{0.49\linewidth}
        \includegraphics[width=\linewidth]{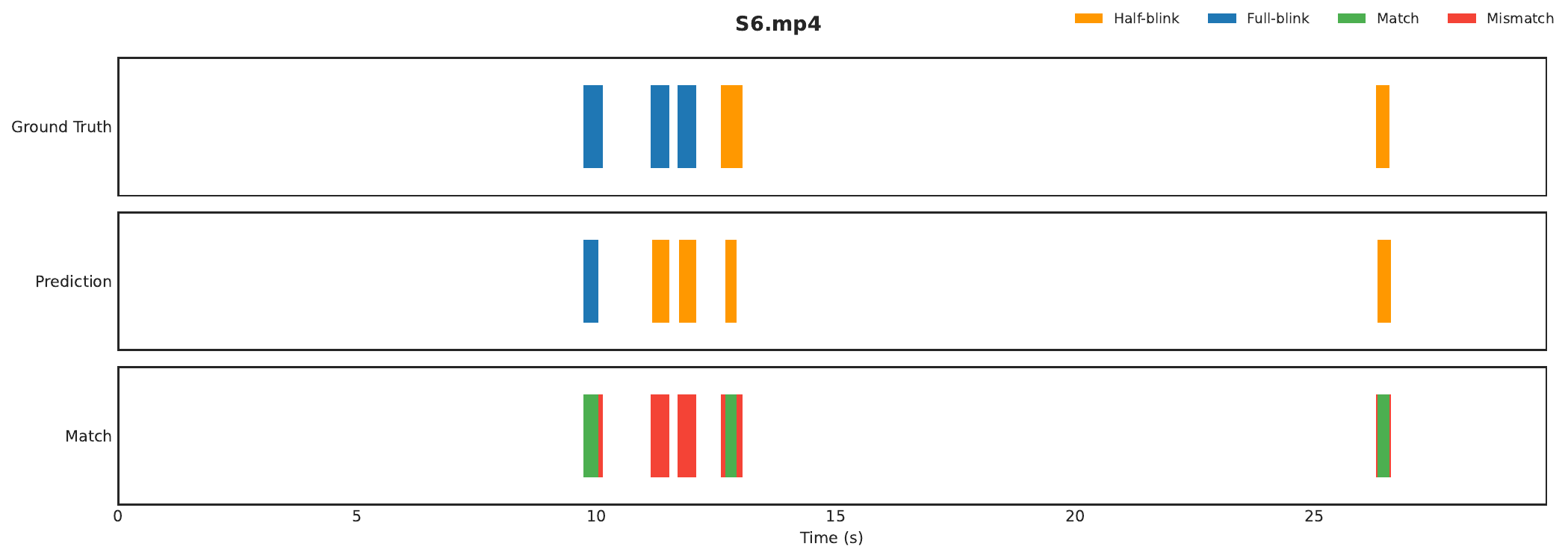}
    \end{subfigure}
    \\[0.5em]
    \begin{subfigure}{0.49\linewidth}
        \includegraphics[width=\linewidth]{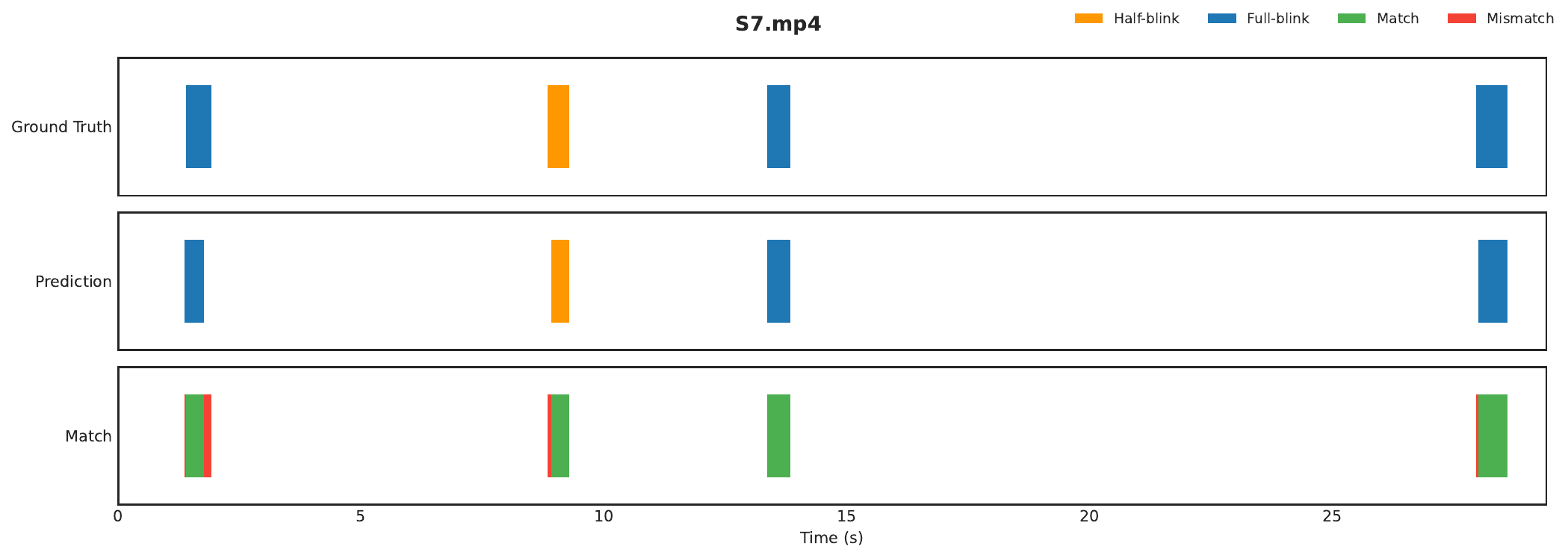}
    \end{subfigure}
    \begin{subfigure}{0.49\linewidth}
        \includegraphics[width=\linewidth]{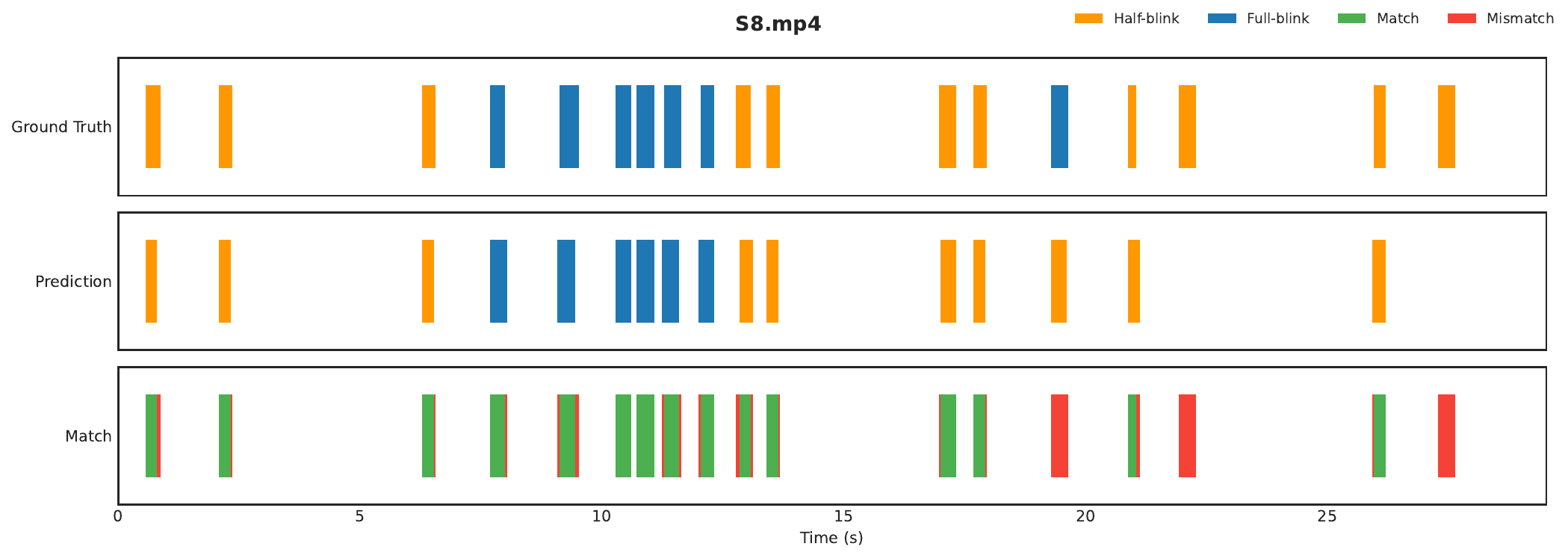}
    \end{subfigure}
    \\[0.5em]
    \begin{subfigure}{0.49\linewidth}
        \includegraphics[width=\linewidth]{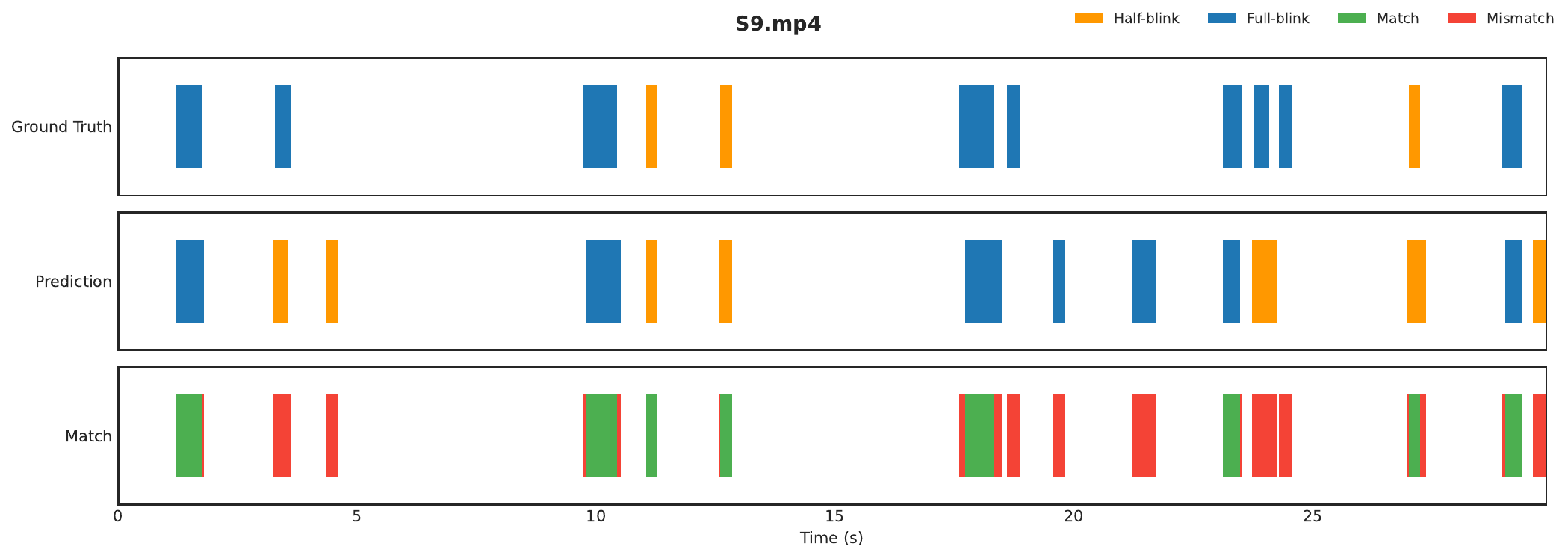}
    \end{subfigure}
    \begin{subfigure}{0.49\linewidth}
        \includegraphics[width=\linewidth]{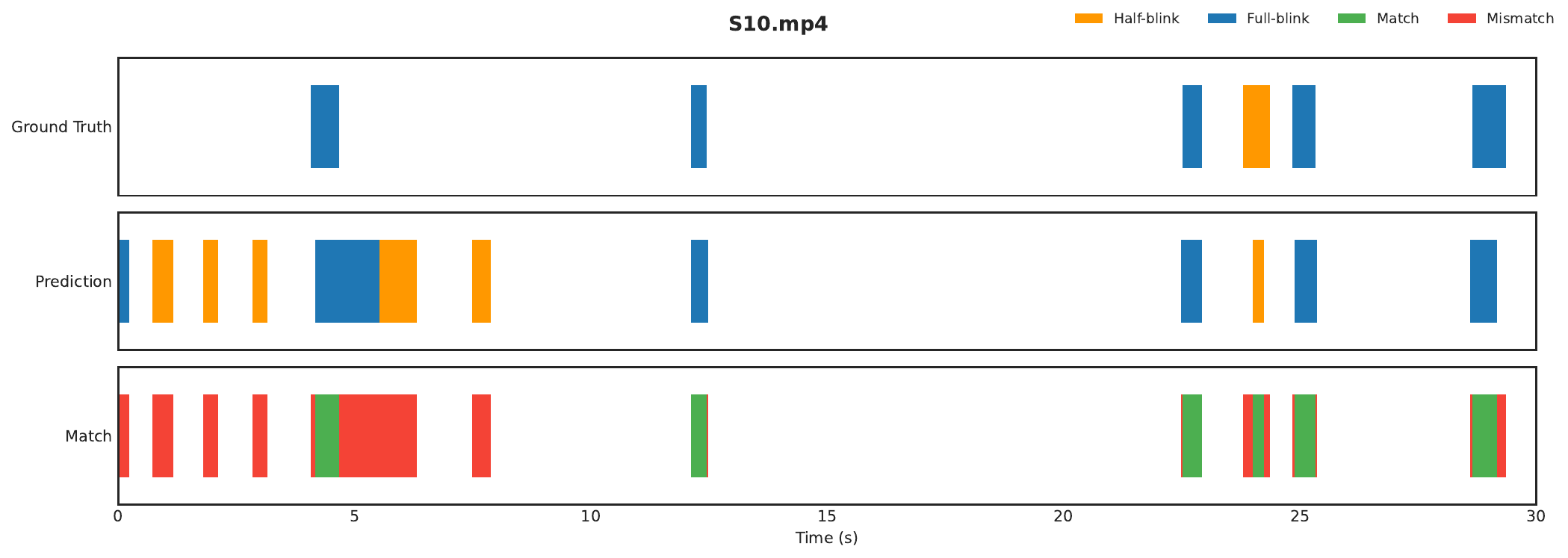}
    \end{subfigure}
    \\[0.5em]
    \begin{subfigure}{0.49\linewidth}
        \includegraphics[width=\linewidth]{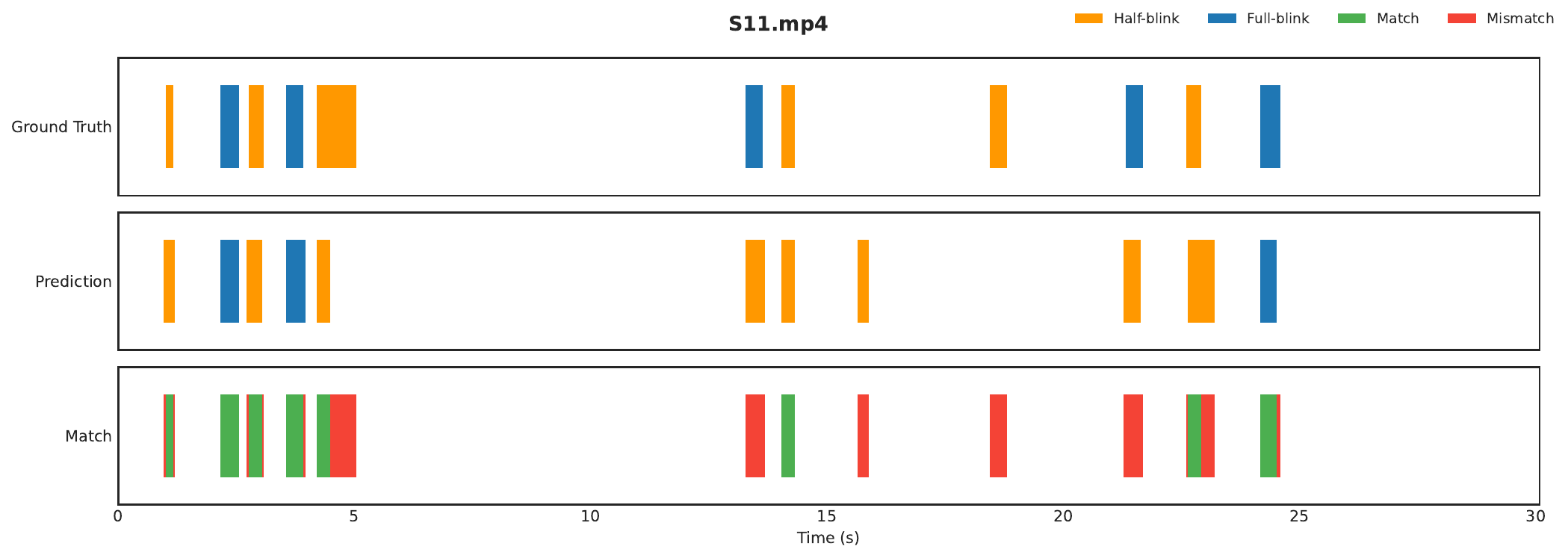}
    \end{subfigure}
    \begin{subfigure}{0.49\linewidth}
        \includegraphics[width=\linewidth]{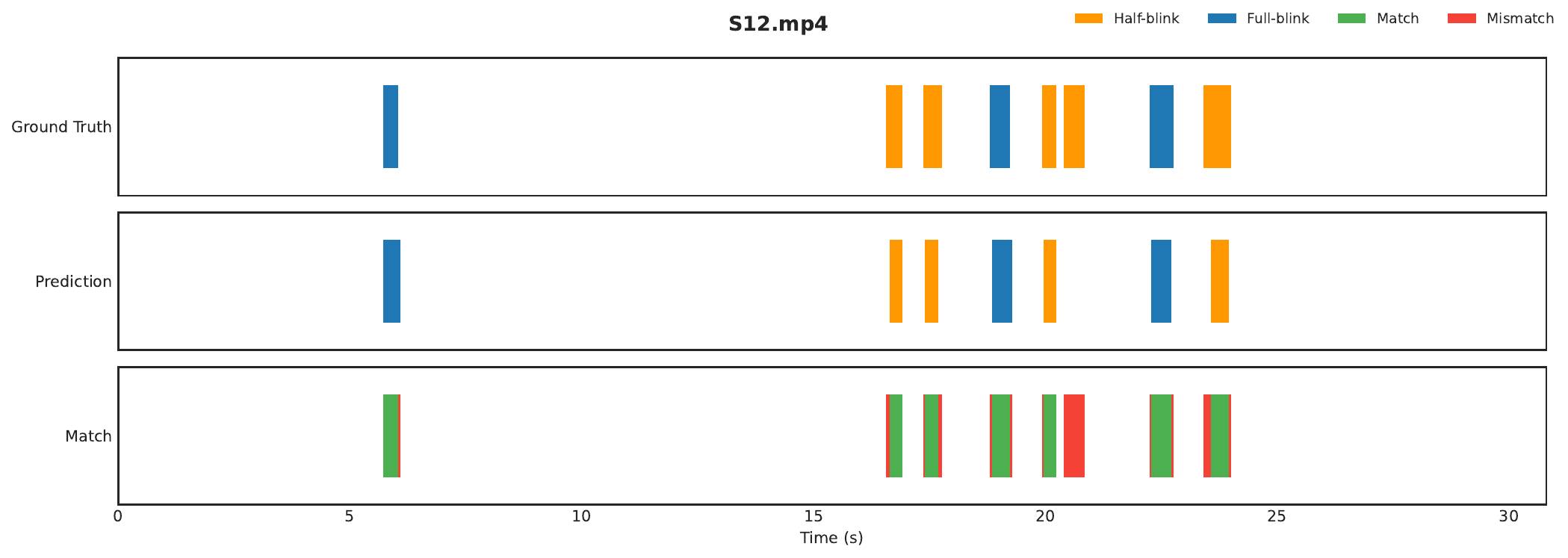}
    \end{subfigure}
    \caption{Qualitative video timelines (S1-S12) for YOLO based eye blink classification method.}
    \label{fig:yolo_qual}
\end{figure*}

\begin{figure*}[htbp]
    \centering
    \begin{subfigure}{0.49\linewidth}
        \includegraphics[width=\linewidth]{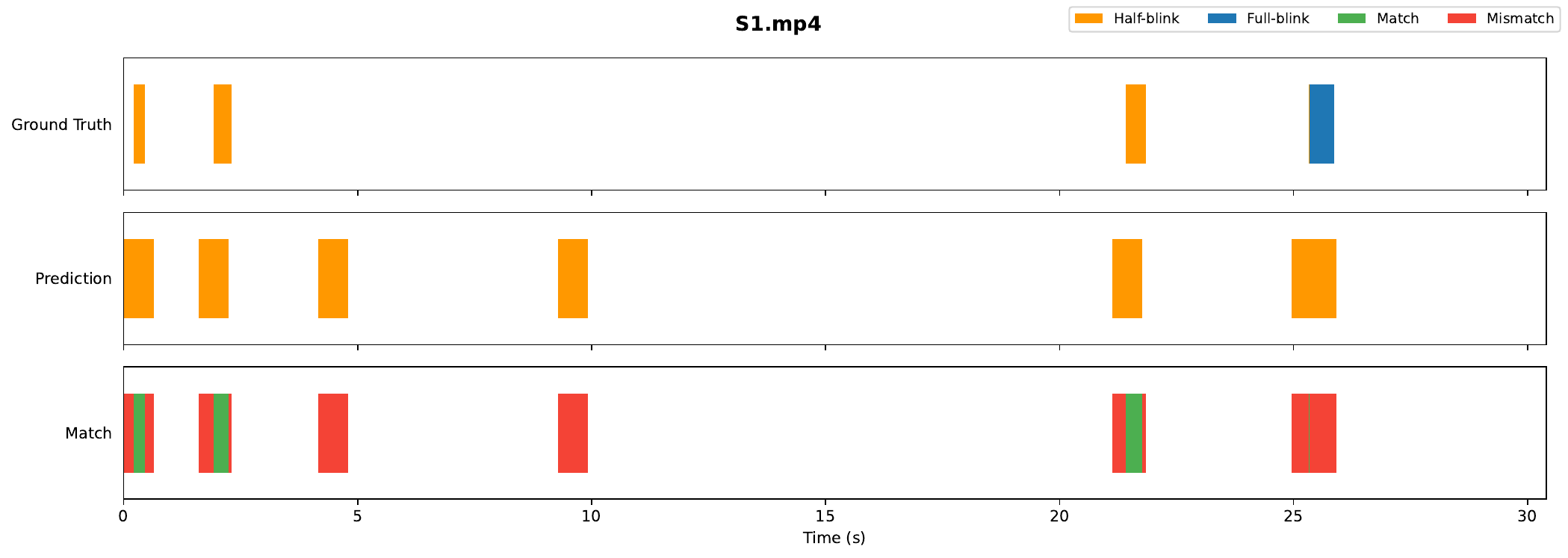}
    \end{subfigure}
    \begin{subfigure}{0.49\linewidth}
        \includegraphics[width=\linewidth]{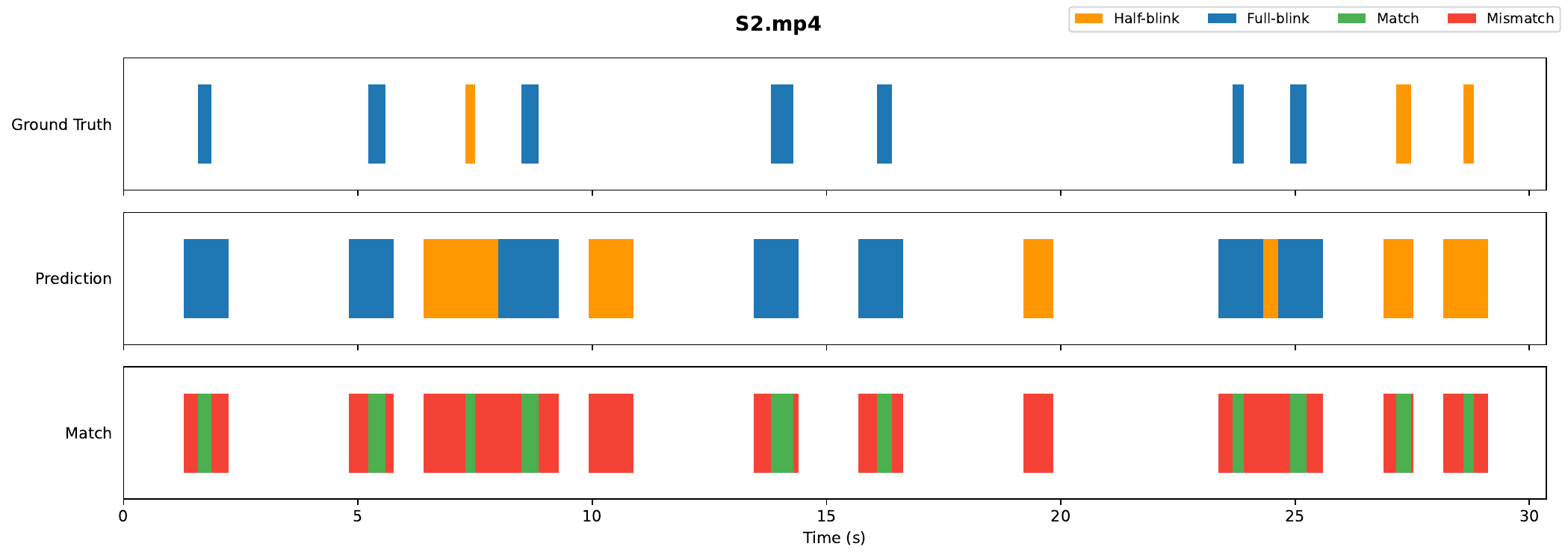}
    \end{subfigure}
    \\[0.5em]
    \begin{subfigure}{0.49\linewidth}
        \includegraphics[width=\linewidth]{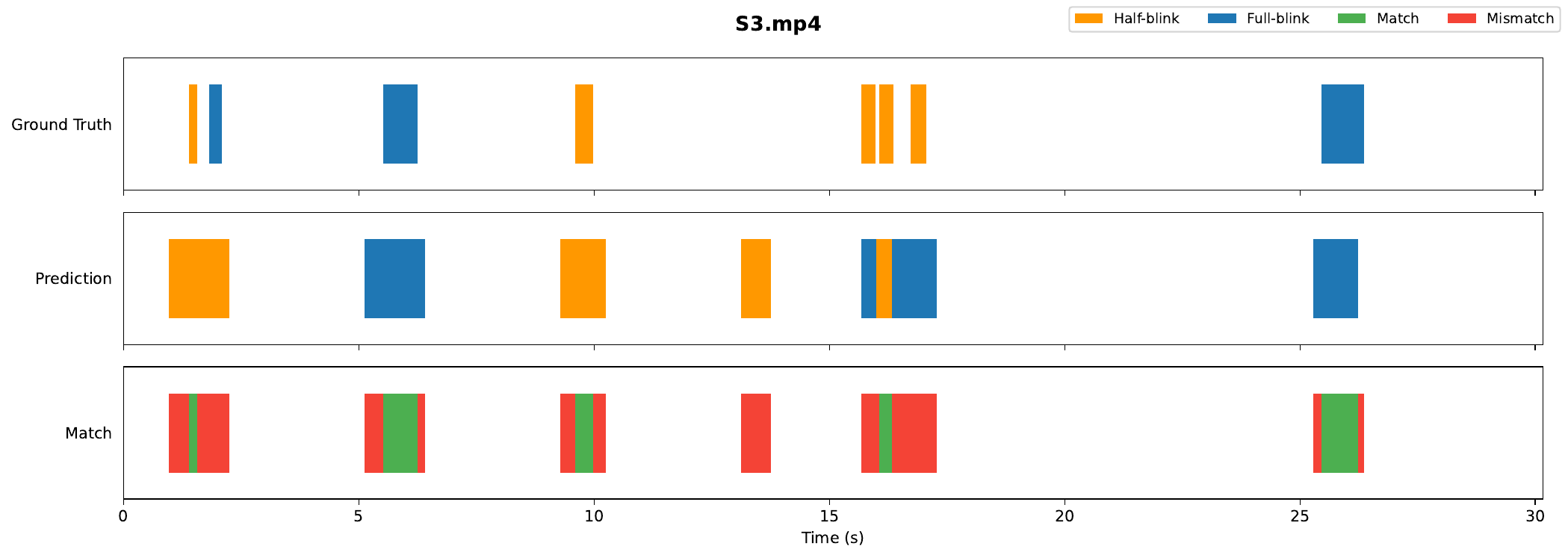}
    \end{subfigure}
    \begin{subfigure}{0.49\linewidth}
        \includegraphics[width=\linewidth]{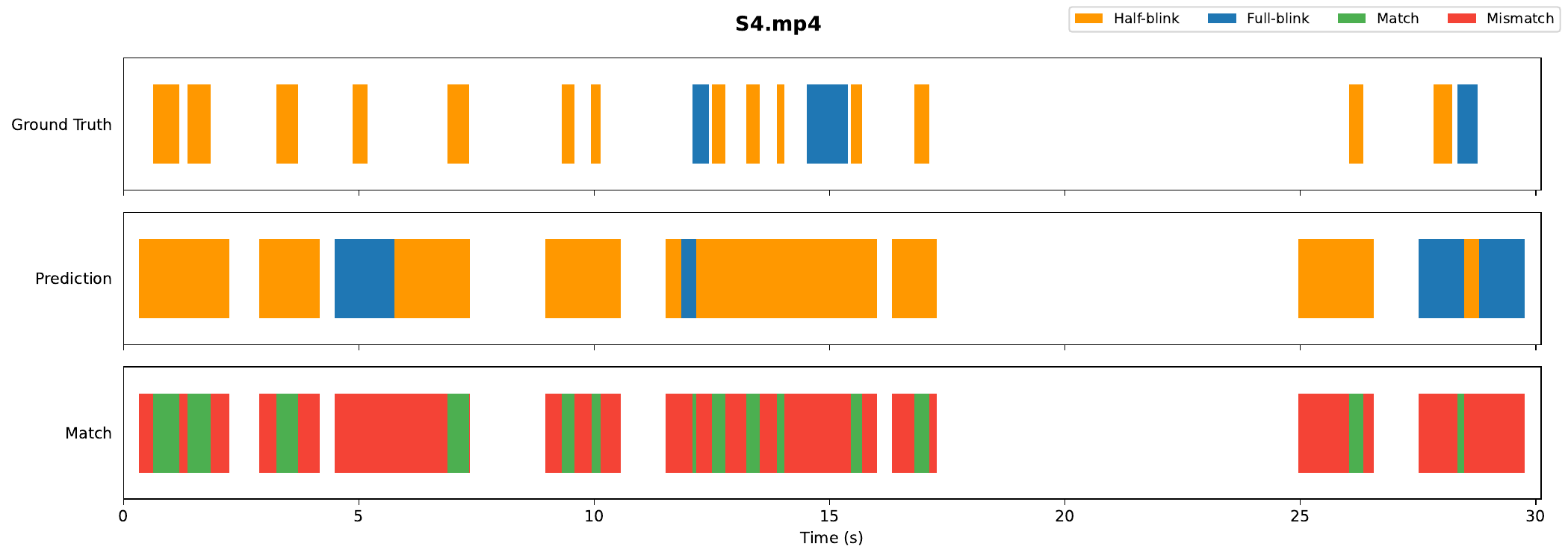}
    \end{subfigure}
    \\[0.5em]
    \begin{subfigure}{0.49\linewidth}
        \includegraphics[width=\linewidth]{images/videomae/qualitative_results/S5_predictions.pdf}
    \end{subfigure}
    \begin{subfigure}{0.49\linewidth}
        \includegraphics[width=\linewidth]{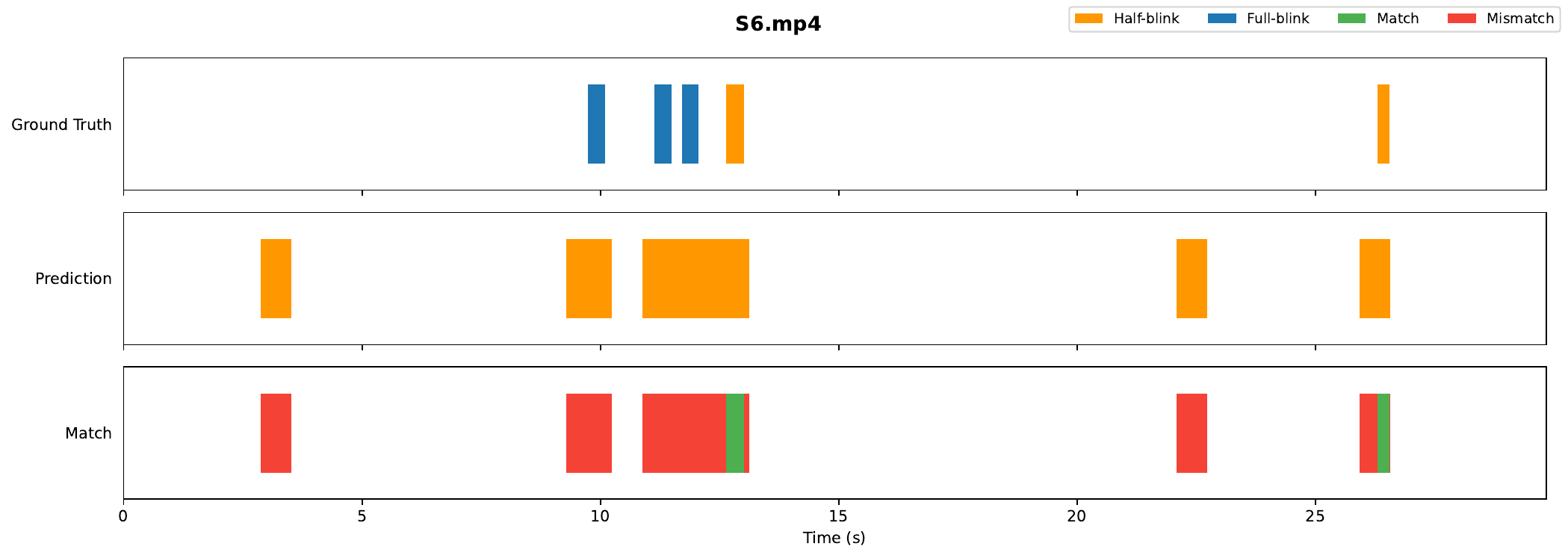}
    \end{subfigure}
    \\[0.5em]
    \begin{subfigure}{0.49\linewidth}
        \includegraphics[width=\linewidth]{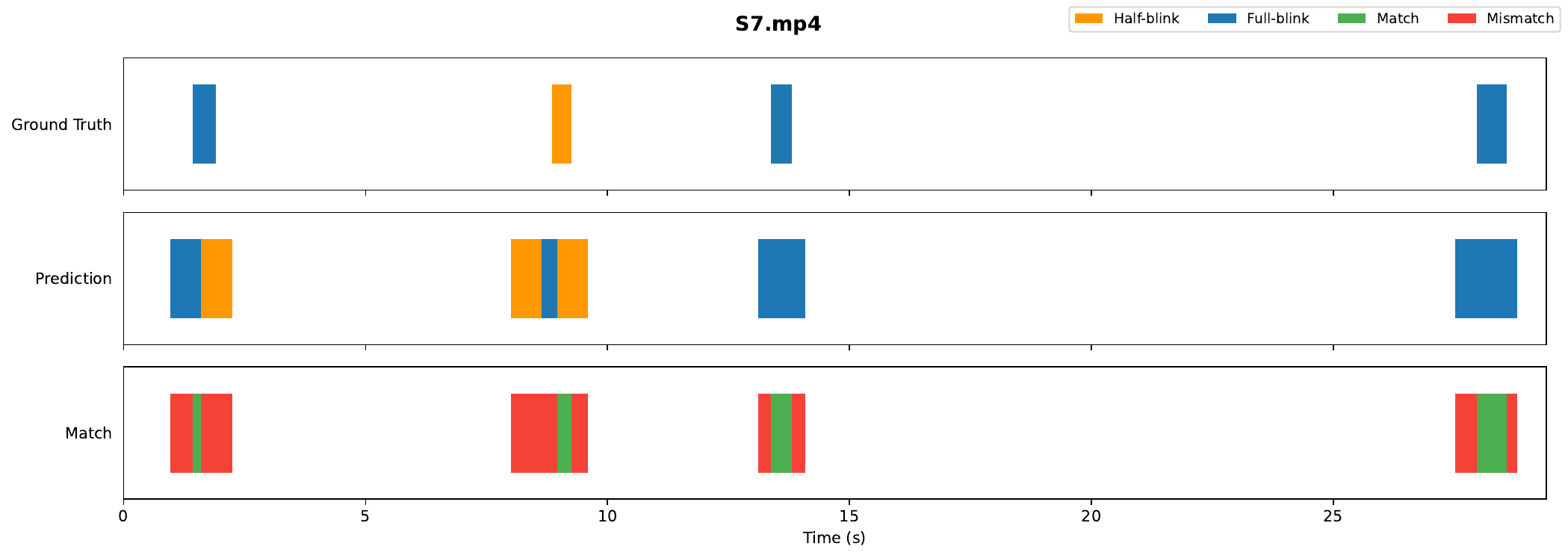}
    \end{subfigure}
    \begin{subfigure}{0.49\linewidth}
        \includegraphics[width=\linewidth]{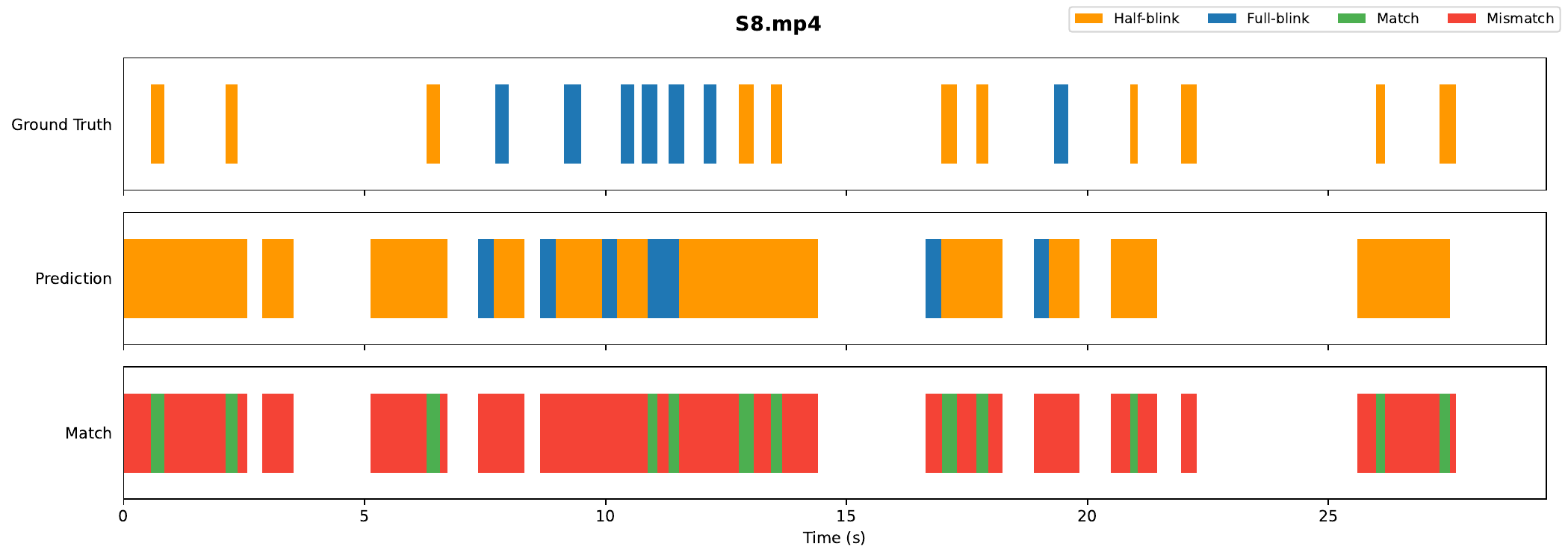}
    \end{subfigure}
    \\[0.5em]
    \begin{subfigure}{0.49\linewidth}
        \includegraphics[width=\linewidth]{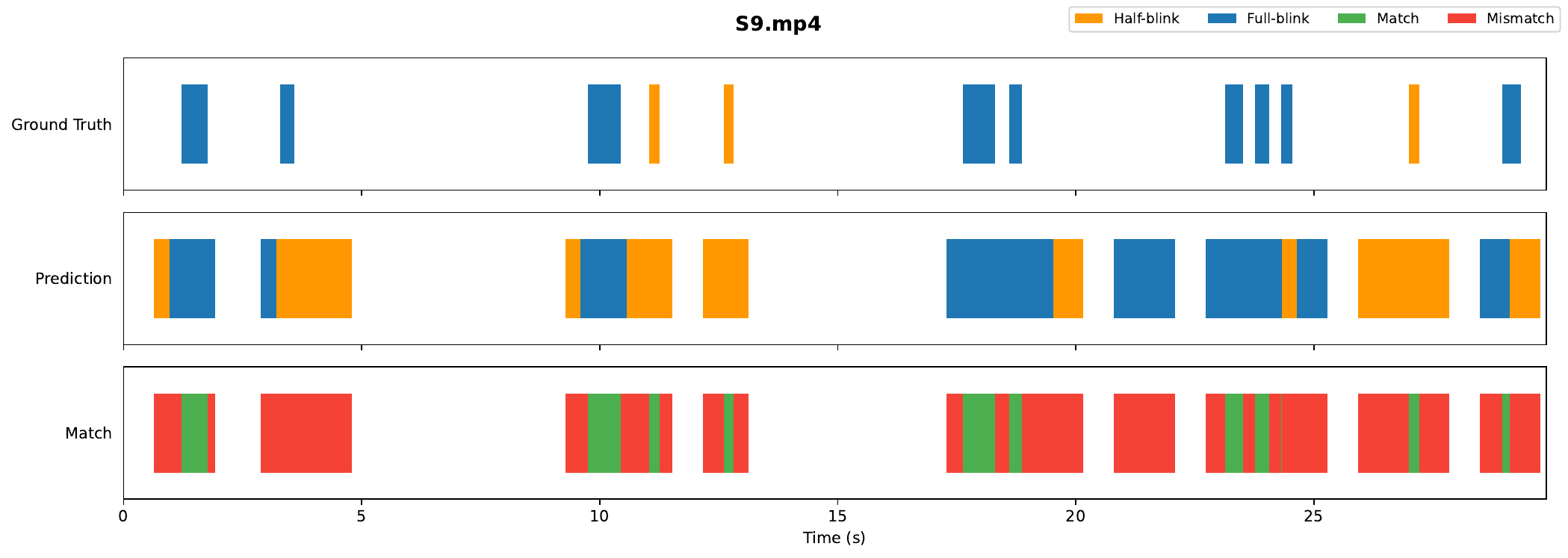}
    \end{subfigure}
    \begin{subfigure}{0.49\linewidth}
        \includegraphics[width=\linewidth]{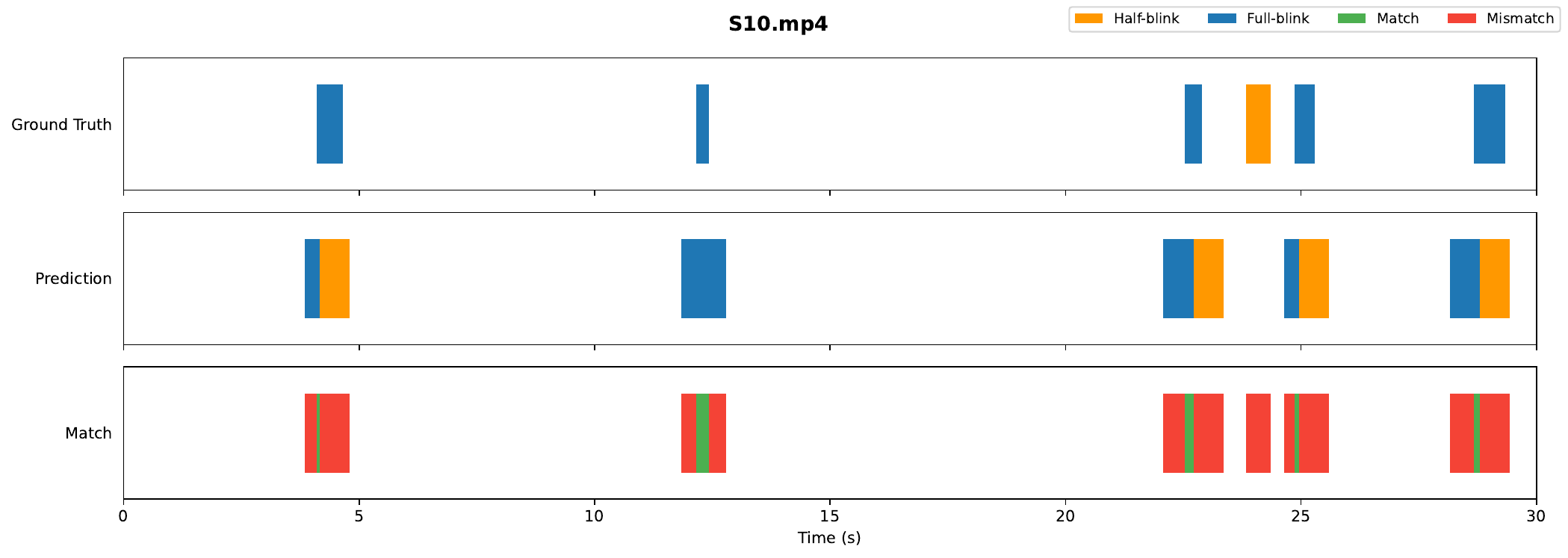}
    \end{subfigure}
    \\[0.5em]
    \begin{subfigure}{0.49\linewidth}
        \includegraphics[width=\linewidth]{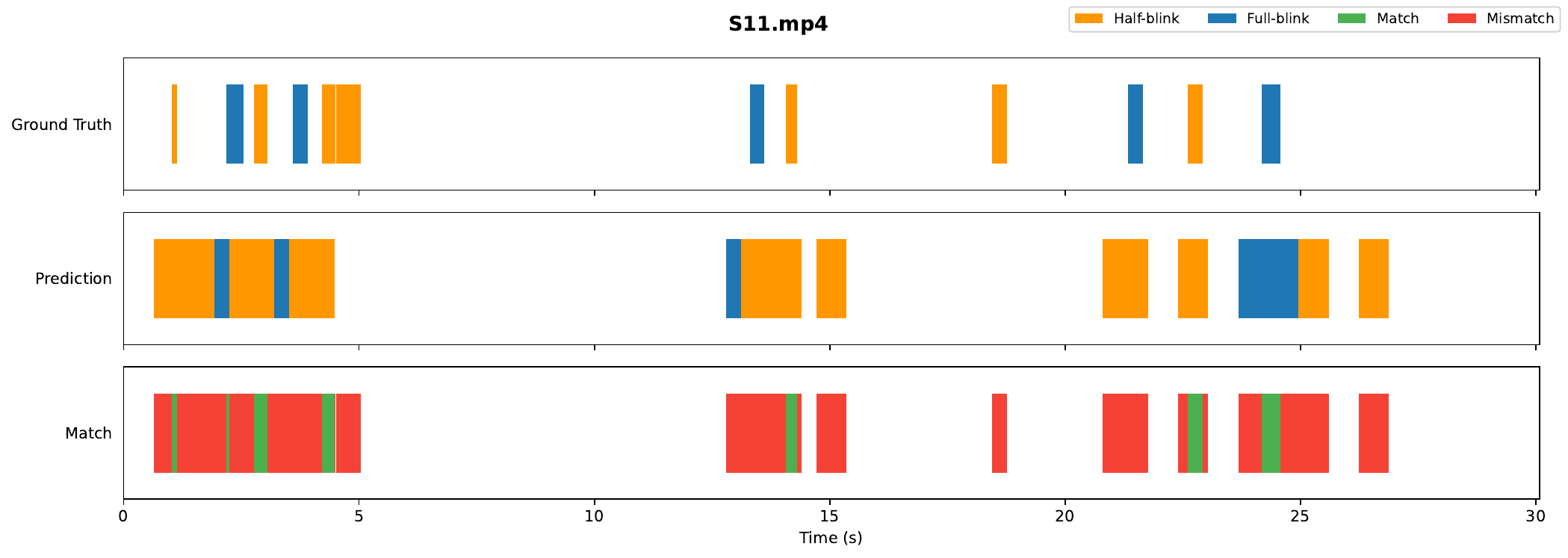}
    \end{subfigure}
    \begin{subfigure}{0.49\linewidth}
        \includegraphics[width=\linewidth]{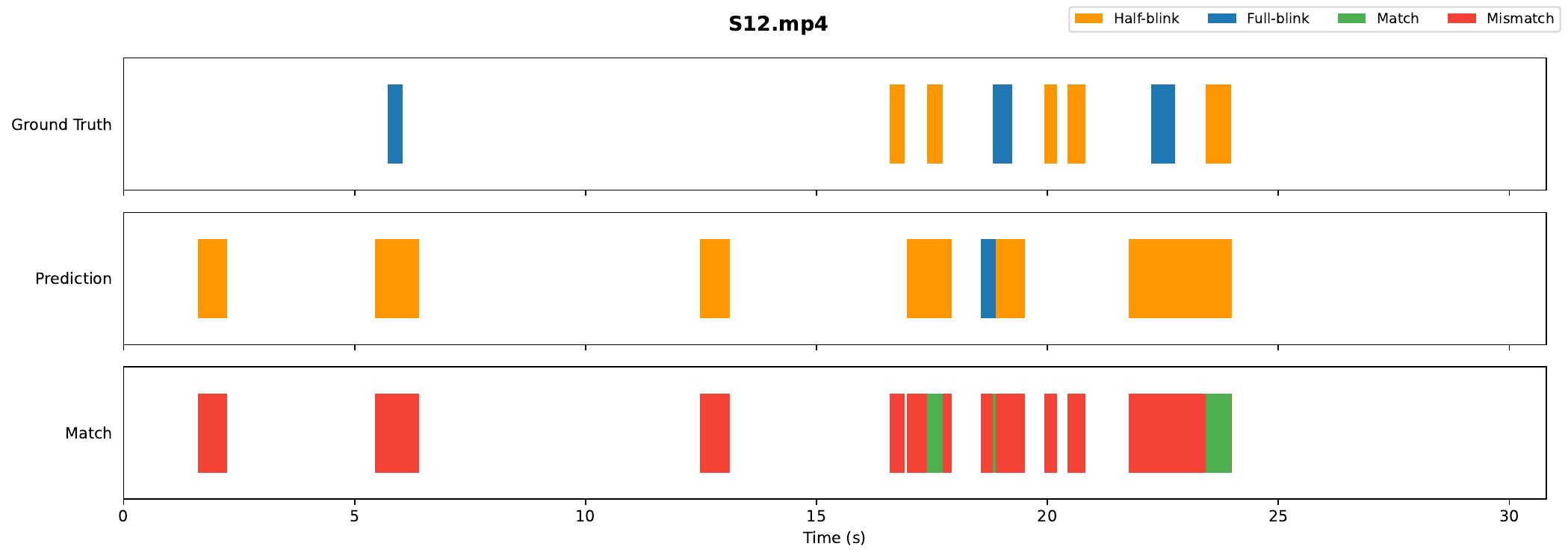}
    \end{subfigure}
    \caption{Qualitative video timelines (S1-S12) for VideoMAE based eye blink classification method following a sliding approach with a window size of 16 frames and stride of 8 frames.}
    \label{fig:videomae_qual}
\end{figure*}

In this section we show the qualitative results obtained from using each method on the original full length dataset videos (see Figures \ref{fig:of_quali}, \ref{fig:yolo_qual} and \ref{fig:videomae_qual}).

\clearpage
\onecolumn

\clearpage

\begin{table*}[htbp]
\scriptsize
\centering
\begin{tabular}{lccccccccccccc}
\toprule
 & & & & & & \multicolumn{2}{c}{start vs GT (frames)} & \multicolumn{2}{c}{end vs GT (frames)} & \multicolumn{2}{c}{Duration vs GT (frames)} & \multicolumn{2}{c}{IoU vs GT} \\
\cmidrule(lr){7-8} \cmidrule(lr){9-10} \cmidrule(lr){11-12} \cmidrule(lr){13-14}
Class & $Ntotal$ & Accuracy & Precision & Recall & F1 & $\mu \pm \sigma$ & Signed & $\mu \pm \sigma$ & Signed & $\mu \pm \sigma$ & Signed & $\mu$ & $\sigma$ \\
\midrule
Nothing    & 48  & 68.8\% & 1.00 & 0.69 & 0.81 & --- & --- & --- & --- & --- & --- & --- & --- \\
Half-blink & 30  & 66.7\% & 0.53 & 0.67 & 0.59 & $1.63 \pm 2.31$ & $+0.83$ & $1.35 \pm 1.35$ & $-0.97$ & $2.60 \pm 2.48$ & $-1.80$ & .66 & .19 \\
Blink      & 54  & 94.4\% & 0.84 & 0.94 & 0.89 & $1.70 \pm 1.21$ & $+1.37$ & $3.50 \pm 2.85$ & $-2.61$ & $4.76 \pm 3.45$ & $-3.98$ & .70 & .16 \\
\midrule
Overall    & 132 & 78.8\% & 0.79 & 0.77 & 0.76 & $1.75 \pm 1.69$ & $+1.18$ & $2.88 \pm 2.59$ & $-2.02$ & $3.92 \pm 3.21$ & $-3.20$ & .68 & .16 \\
\bottomrule
\end{tabular}
\caption{Annotator accuracy relative to ground truth: classification performance and temporal boundary error by class. $Ntotal$ is the total number of classified clips. Mean is represented by $\mu$ and standard deviation by $\sigma$.}
\label{tab:inter_study2}
\end{table*}

\begin{figure*}[htbp]
    \centering
    \begin{subfigure}{0.49\linewidth}
        \includegraphics[width=\linewidth]{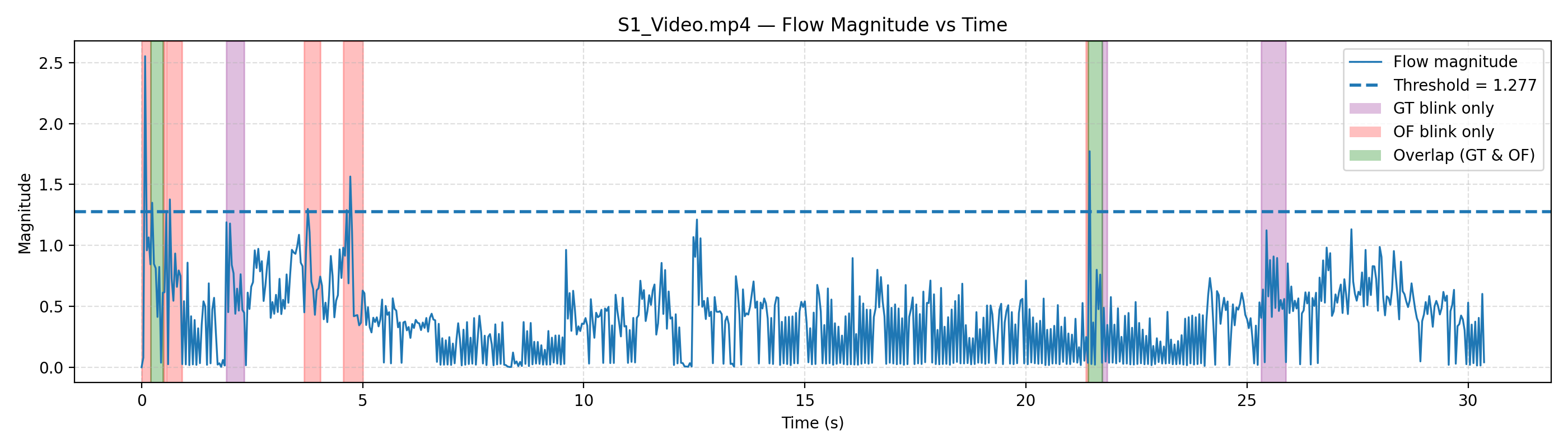}
    \end{subfigure}
    \begin{subfigure}{0.49\linewidth}
        \includegraphics[width=\linewidth]{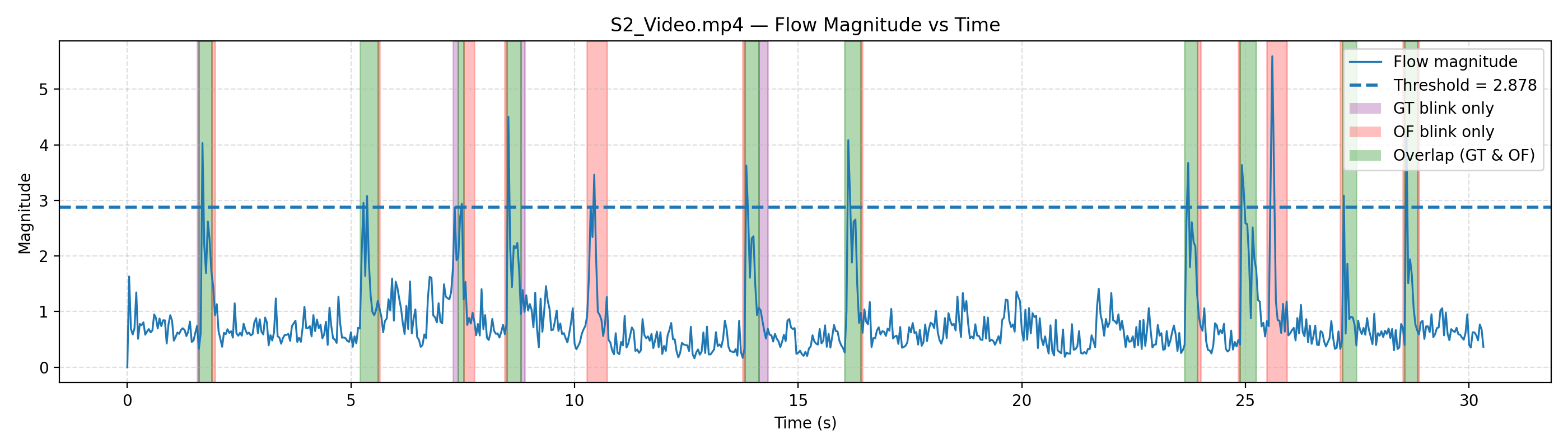}
    \end{subfigure}
    \\[0.5em]
    \begin{subfigure}{0.49\linewidth}
        \includegraphics[width=\linewidth]{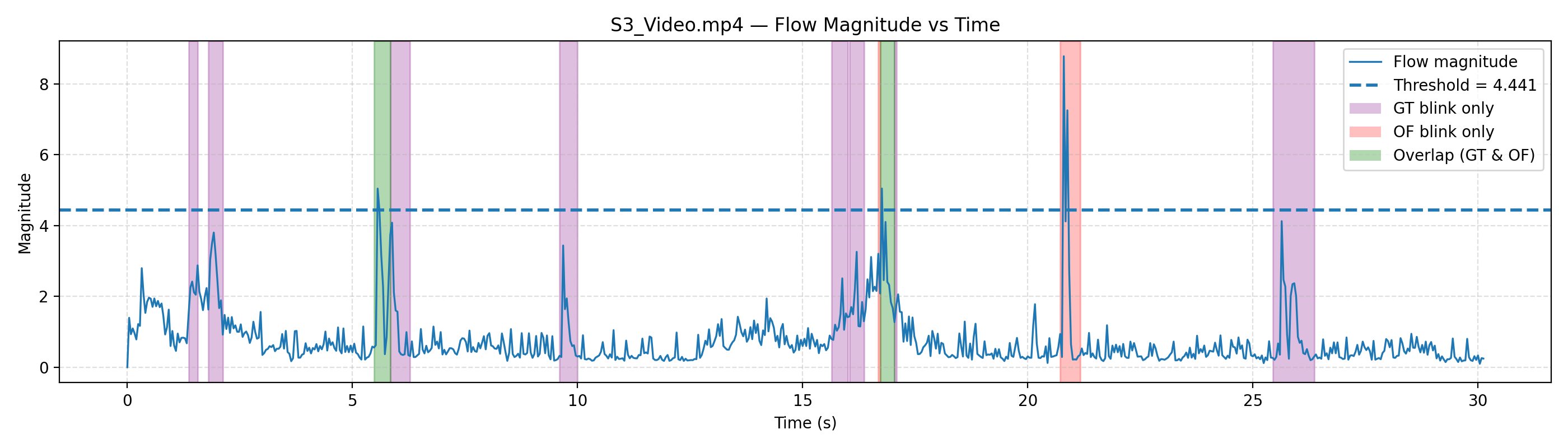}
    \end{subfigure}
    \begin{subfigure}{0.49\linewidth}
        \includegraphics[width=\linewidth]{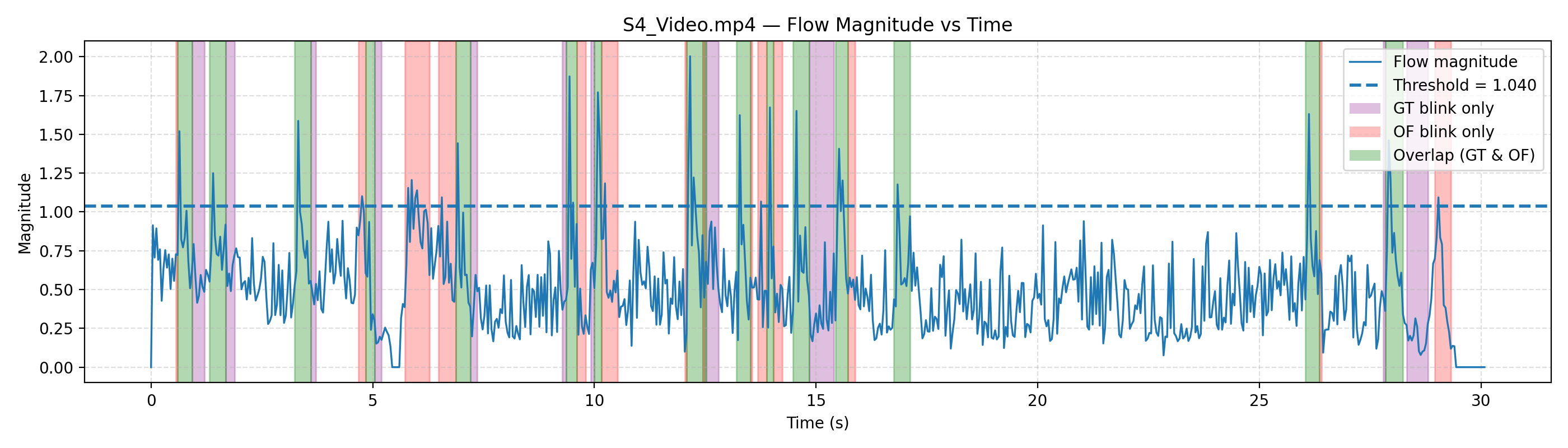}
    \end{subfigure}
    \\[0.5em]
    \begin{subfigure}{0.49\linewidth}
        \includegraphics[width=\linewidth]{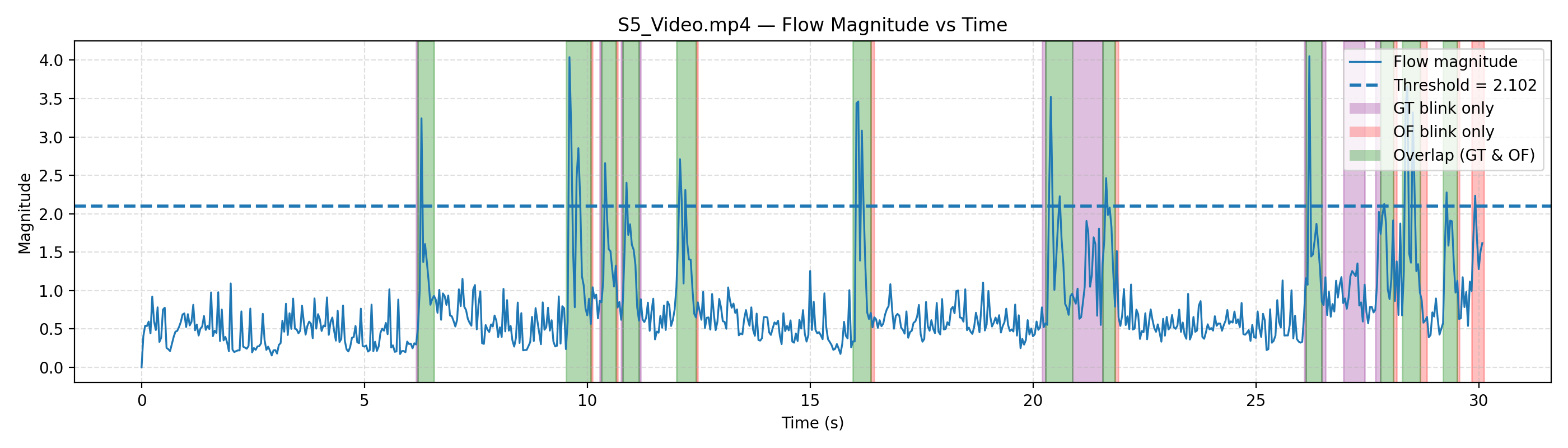}
    \end{subfigure}
    \begin{subfigure}{0.49\linewidth}
        \includegraphics[width=\linewidth]{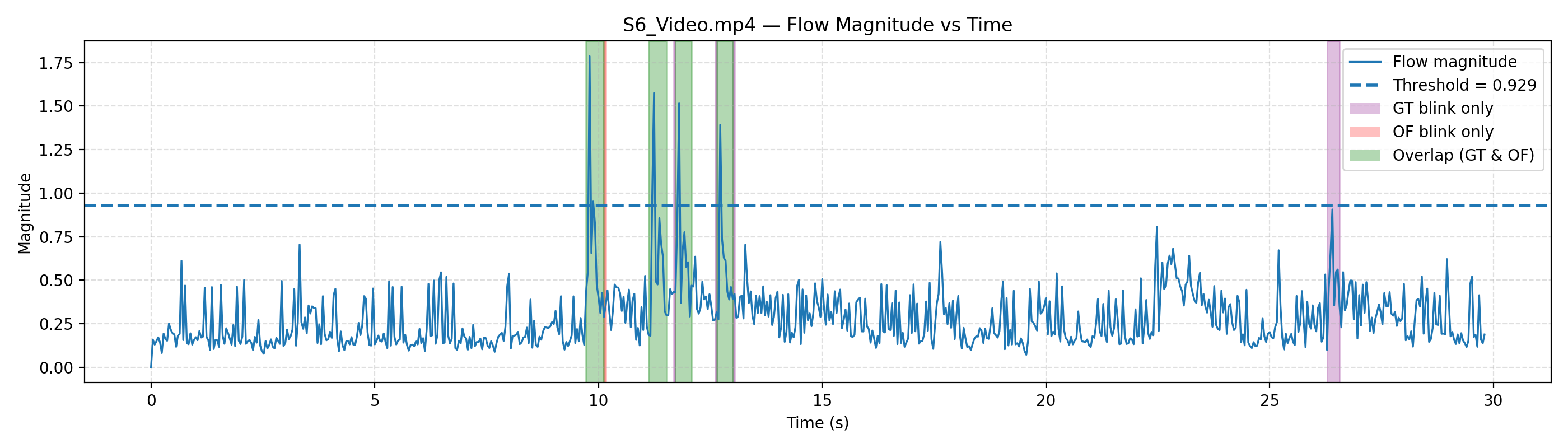}
    \end{subfigure}
    \\[0.5em]
    \begin{subfigure}{0.49\linewidth}
        \includegraphics[width=\linewidth]{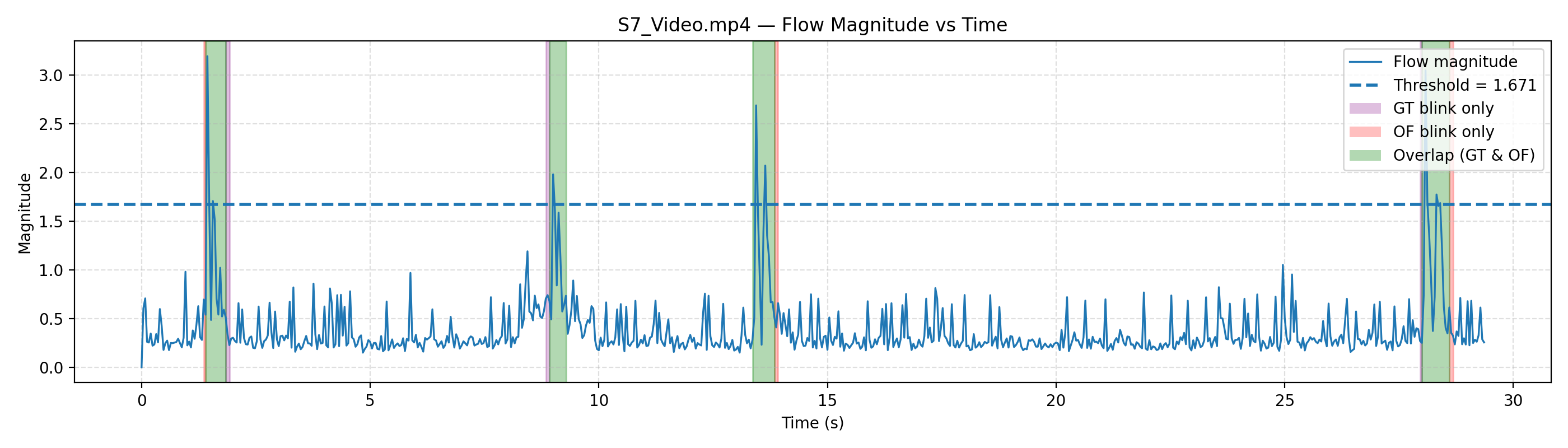}
    \end{subfigure}
    \begin{subfigure}{0.49\linewidth}
        \includegraphics[width=\linewidth]{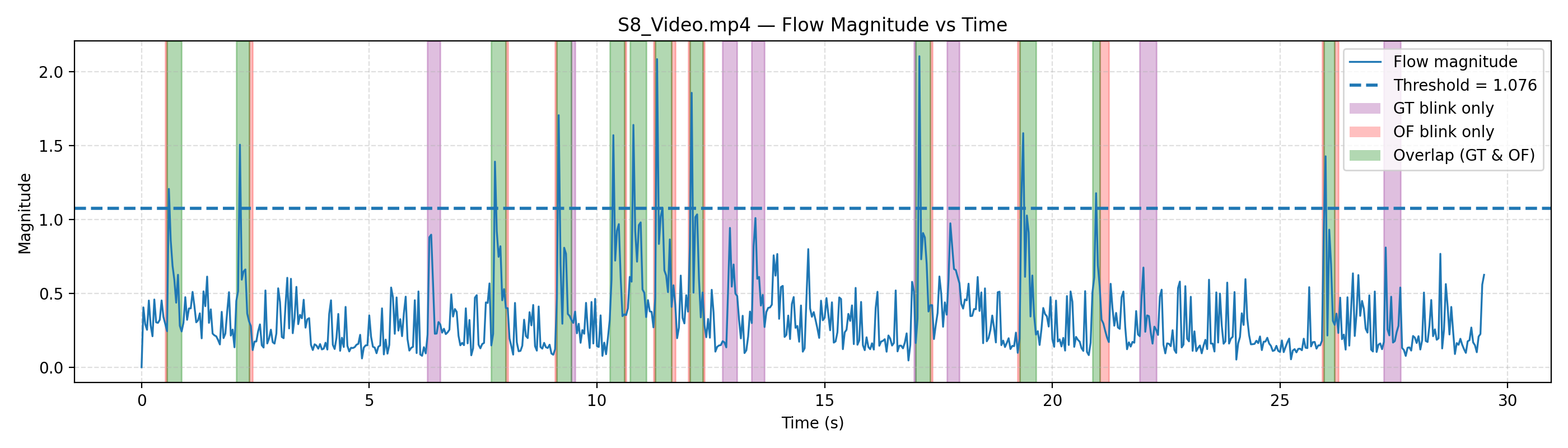}
    \end{subfigure}
    \\[0.5em]
    \begin{subfigure}{0.49\linewidth}
        \includegraphics[width=\linewidth]{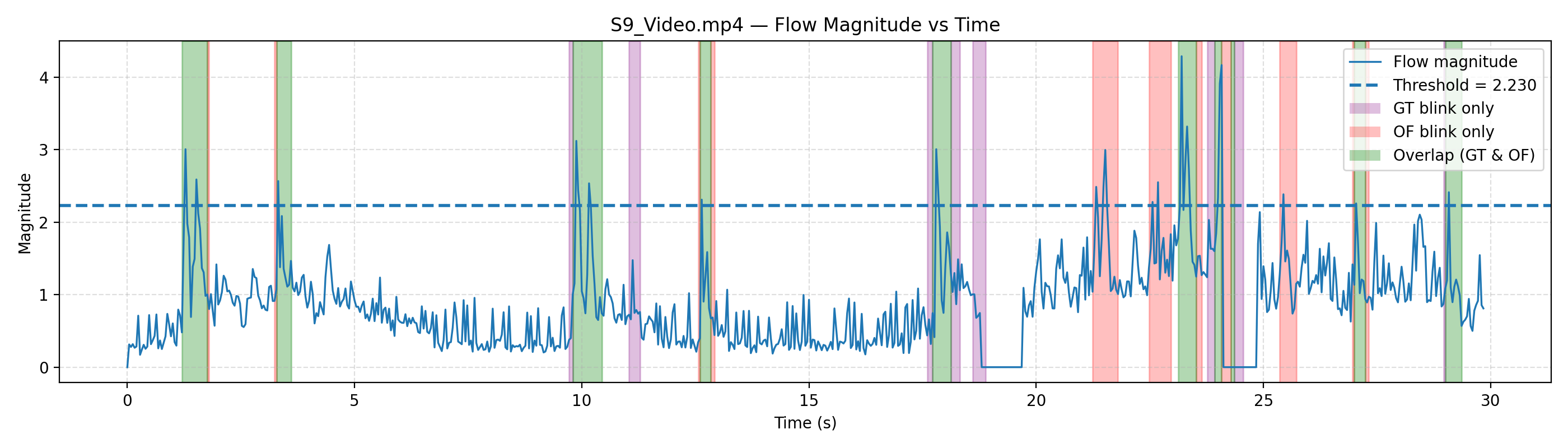}
    \end{subfigure}
    \begin{subfigure}{0.49\linewidth}
        \includegraphics[width=\linewidth]{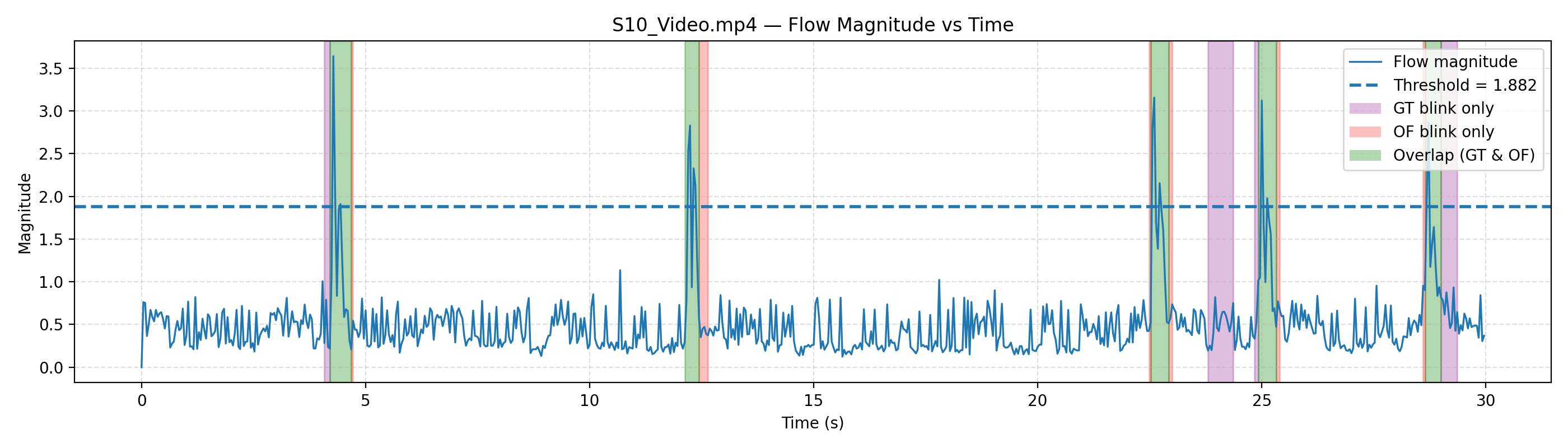}
    \end{subfigure}
    \\[0.5em]
    \begin{subfigure}{0.49\linewidth}
        \includegraphics[width=\linewidth]{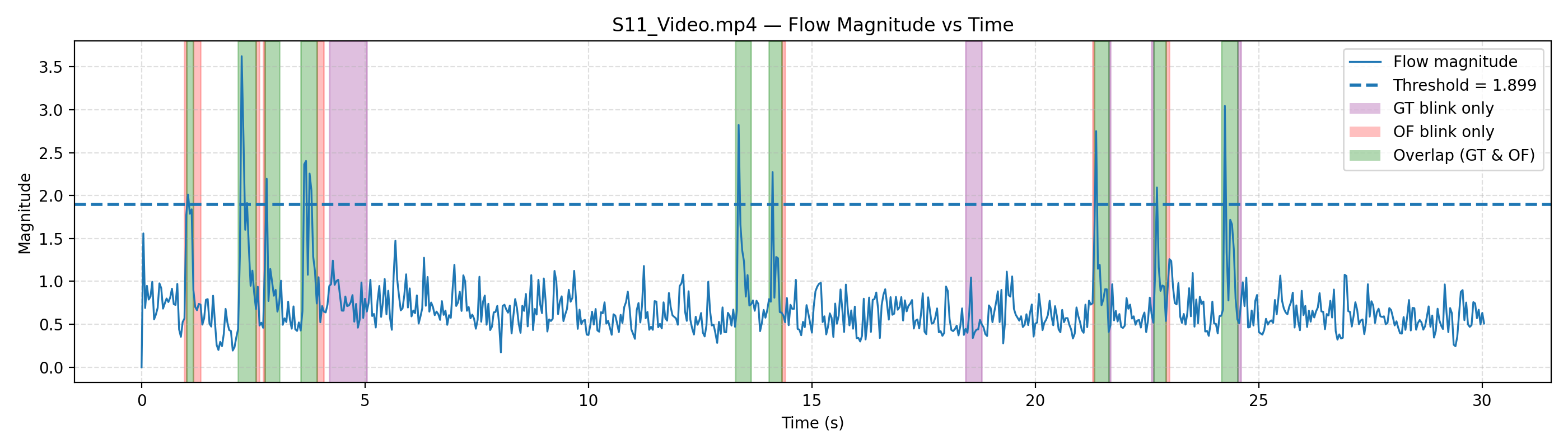}
    \end{subfigure}
    \begin{subfigure}{0.49\linewidth}
        \includegraphics[width=\linewidth]{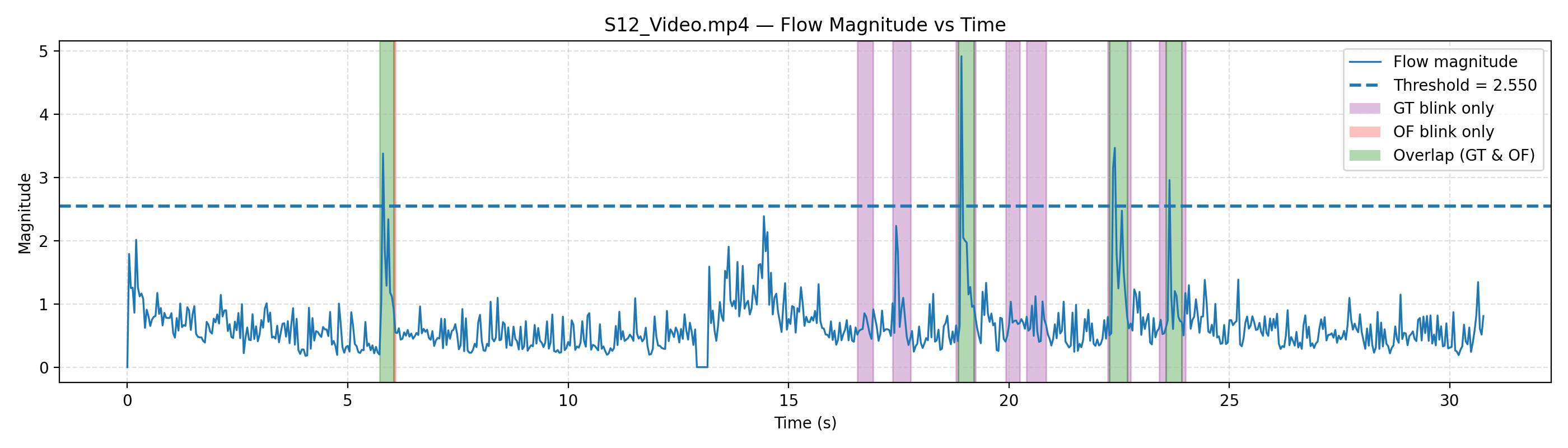}
    \end{subfigure}
    \caption{Qualitative video timelines (S1-S12) for Optical Flow based eyelid closure detection method.}
    \label{fig:sample_of_gt_pred_perf}
\end{figure*}

